\documentclass{ieeeaccess}
\usepackage{cite}
\usepackage{amsmath,amssymb,amsfonts}
\usepackage{algorithmic}
\usepackage{graphicx}
\usepackage{textcomp}

\usepackage{bm}
\makeatletter
\AtBeginDocument{\DeclareMathVersion{bold}
\SetSymbolFont{operators}{bold}{T1}{times}{b}{n}
\SetSymbolFont{NewLetters}{bold}{T1}{times}{b}{it}
\SetMathAlphabet{\mathrm}{bold}{T1}{times}{b}{n}
\SetMathAlphabet{\mathit}{bold}{T1}{times}{b}{it}
\SetMathAlphabet{\mathbf}{bold}{T1}{times}{b}{n}
\SetMathAlphabet{\mathtt}{bold}{OT1}{pcr}{b}{n}
\SetSymbolFont{symbols}{bold}{OMS}{cmsy}{b}{n}
\renewcommand\boldmath{\@nomath\boldmath\mathversion{bold}}}
\makeatother

\def\BibTeX{{\rm B\kern-.05em{\sc i\kern-.025em b}\kern-.08em
    T\kern-.1667em\lower.7ex\hbox{E}\kern-.125emX}}

\usepackage{makecell}
\usepackage{stackengine}
\usepackage{multirow}
\usepackage{caption}
\usepackage{subfig}
\usepackage{colortbl}
\usepackage{amssymb}
\usepackage{diagbox}
\usepackage{xcolor}

\begin{document}
\history{Date of publication xxxx 00, 0000, date of current version xxxx 00, 0000.}
\doi{10.1109/ACCESS.2024.0429000}

\title{Predicting Multiple Clinical Outcomes Related to Functional Recovery and Social Isolation Among Older Adults After Lower-Limb Fracture or Hip Replacement}
\author{\uppercase{Santosh Ray}\authorrefmark{1},
\uppercase{Pratik K. Mishra}\authorrefmark{2,3},
\uppercase{Ali Abedi}\authorrefmark{4,5},
\uppercase{Charlene H. Chu}\authorrefmark{3,5},
\uppercase{Amir Ahmad}\authorrefmark{6}, AND 
\uppercase{Shehroz S. Khan} \authorrefmark{7}}

\address[1]{Faculty of Information Technology, Liwa University, Abu Dhabi, United Arab Emirates}
\address[2]{Institute of Biomedical Engineering, University of Toronto, Toronto, ON, Canada}
\address[3]{KITE Research Institute - Toronto Rehabilitation Institute, University Health Network, Toronto, Canada}
\address[4]{Peter Munk Cardiac Centre, University Health Network, Toronto, Canada}
\address[5]{Lawrence Bloomberg Faculty of Nursing, University of Toronto, Toronto, Canada}
\address[6]{College of Information Technology, United Arab Emirates University, Al Ain, United Arab Emirates}
\address[7]{College of Engineering and Technology, American University of the Middle East, Egaila, 54200, Kuwait}
\tfootnote{This study is funded by Liwa University, United Arab Emirates (Grant: IRG-ENG-003-2025).}

\markboth
{Author \headeretal: Preparation of Papers for IEEE TRANSACTIONS and JOURNALS}
{Author \headeretal: Preparation of Papers for IEEE TRANSACTIONS and JOURNALS}

\corresp{Corresponding author: Santosh Ray (e-mail: santosh.ray@lu.ac.ae).}

\begin{abstract}
Older adults recovering after lower-limb fracture or hip replacement may experience complex recovery trajectories. Most of the time, these clinical aspects are studied in isolation, masking their joint impact on recovery. This study used the MAISON-LLF dataset, which contains multimodal sensor and clinical assessment data from 18 older adults recovering in the community after lower-limb fracture or hip replacement. Participants were monitored for up to eight weeks, corresponding to a maximum of 1,008 participant-days of sensor monitoring. Forty-six daily features were extracted from indoor motion, acceleration, step count, heart rate, out-of-home mobility, and sleep data. Five clinical outcomes were assessed every two weeks: the Social Isolation Scale, Oxford Hip Score, Oxford Knee Score, Timed Up and Go test, and 30-second Chair Stand test. We utilize an inherent relationship between multi-modal sensor data and different clinical scores and formulate it as a multi-output regression problem. We tested various machine learning and deep learning single- and multi-output regression algorithms to predict these scores simultaneously. The results showed that predicting clinical scores jointly was better than separately. The tabular DL multi-output regressor, NODE, gave a remarkable performance of MSE=3.96 and MAE=1.02 in comparison to other multi- and single-output regressors. The SHAP feature analysis further showed the importance of including multimodal sensors to provide a good estimate of patients' recovery trajectory. This work may support the simultaneous assessment of functional recovery and social engagement among community-dwelling older adults and ultimately help improve their care and quality of life.
\end{abstract}


\titlepgskip=-21pt

\maketitle

\section{Introduction} \label{sec_introduction}
\label{sec:intro}

With advancements in modern medicine, lower-limb surgery patients are often discharged from hospitals quickly to recover in the community to reduce infection risks and healthcare costs \cite{ishaku2025enhanced}. Recovery after lower-limb fracture or hip replacement can be complex and may involve limitations in mobility and social engagement within the community \cite{singh2025rehabilitation, mohammed2025psychosocial}. Regular assessment of functional and social engagement is mostly not covered by the primary care services, unless the patient meets the designated clinician. Therefore, recovery patterns, functional mobility, and social isolation of these patients are rarely studied in tandem.

Previous studies show that mobility \cite{shear2025predicting}, sleep \cite{rao2026management}, and other physiological indicators (e.g., heart rate) \cite{mangione2005can} are separately correlated with measures of functional mobility and social isolation in post limb-fracture patients. Clinical literature also suggests that there are correlations between functional mobility and social isolation among older adults  \cite{bevilacqua2021association, mandl2024effect}. 
Multimodal sensors provide opportunities to monitor and log various health indicators that are associated with functional mobility and social isolation, such as indoor motion, outdoor movements, sleep quality, heart rate, number of steps, and other physiological data. When studied together, these unique data can provide an overall profile of a patient that can predict the degree of functional decline and social isolation. 

The MAISON-LLF dataset was collected using the cloud-based Multimodal Artificial Intelligence-based Sensor platform for Older iNdividuals (MAISON) \cite{abedi2022maison,abedi2025multimodal}. It includes longitudinal wearable, ambient, sleep, mobility, and clinical assessment data from older adults recovering in the community after lower-limb fracture or hip replacement. Clinical outcomes included the Oxford Hip Score (OHS), Oxford Knee Score (OKS), Timed Up and Go (TUG), 30-second Chair Stand, and Social Isolation Scale (SIS). The MAISON-LLF data is made publicly available by the authors \cite{maison-llf}. Prior work on using sub-samples of MAISON-LLF shows a correlation between sensor data collected for mobility and sleep and functional decline and social isolation \cite{dayyani2024correlations}. In another work, different clinical assessments were predicted separately with low MAE, highlighting the predictive ability of the dataset \cite{maison-llf}. From clinical literature, it is known that different clinical assessments, e.g., OHS, OKS, TUG, Chair Stand and Social Isolation, are interrelated as mobility and social engagement impact the overall recovery in this population. However,  there exists no study that attempts to predict multiple clinical assessments on functional recovery and social isolation simultaneously to harness further clinical value from this dataset. The positive outcomes can have significant implications for improving the care and quality of life of patients, reducing readmissions, and lessening caregiver burden. Furthermore, it could provide an alarm ahead of time to the caregivers/clinicians to take preventive measures.

We formulate the problem of predicting multiple clinical scores from multimodal sensor data collected from the MAISON platform as a multi-output regression problem. This formulation enables training joint models to harness the interrelation between these clinical outcomes in predicting them simultaneously and provides a multidimensional recovery trajectory overview in this population.  
The key contributions of this work are as follows:
\begin{itemize}
    \item We formulate the multimodal sensor data collected from post lower-limb surgery as a multi-output regression problem to be able to predict five clinical measures simultaneously (OHS, OKS, TUG, Chair Stand and SIS). We further show that predicting multiple clinical measures can yield low MAE in comparison to predicting them separately.
    \item We investigated the role of tabular deep learning models on the MAISON-LLF dataset to establish their superior performance in comparison to the traditional machine learning multi-output regression methods.
    \item Feature selection and SHAP analysis further show that the multimodal sensor approach is useful, as data from different sensors appear in the top features list.
\end{itemize}

\section{Related Work}\label{sec:related-work}
This section reviews prior work on predicting social isolation and functional recovery in older adults using multimodal sensor data and machine learning (ML), followed by studies on multi-output regression and multi-task learning for health-related and sensor-based prediction tasks. It then summarizes the remaining gaps in the field and situates our work to 
address them.

\subsection{Predicting Social Isolation and Functional Recovery in Older Adults}
\label{sec:related-work-prediction}
Sensor-based monitoring has increasingly been used to capture behavioural, social, and functional changes in older adults outside clinical environments. Early studies focused on loneliness, social isolation, and social interaction patterns using passive sensing and smart-home technologies. Austin et al. used wireless motion sensors, contact sensors, telephone monitoring, computer-use monitoring, and walking-speed measures and trained a logistic regression classifier to estimate loneliness status in older adults \cite{austin2016smart}. Martinez et al. used communication, mobility, and activity features collected through a mobile application and passive infrared (PIR) sensing, and applied a decision tree model with the synthetic minority oversampling technique (SMOTE) to classify social isolation risk \cite{martinez2020automatic}. Goonawardene et al. used non-intrusive motion and door-contact sensors to extract going-out behavior, daytime napping, and room-use patterns associated with social isolation dimensions \cite{goonawardene2017sensor}. These studies established that daily behavioral traces can provide objective indicators of psychosocial states in older adults. The scoping review by Khan et al. \cite{khan2023sensor} identified very few  studies tackling social isolation using sensors. They found that data from motion sensors and actigraphy were commonly used and correlated with self-report measures in developing objective SI assessments. They reported that variability exists in defining social isolation, feature extraction, use of sensors, and self-report assessments.

Other studies inferred social connectedness indirectly through home visits and in-home activity patterns. Hu et al. used ambient and wearable sensors to detect home visits among older adults living alone, treating visit occurrence as a proxy for social contact \cite{hu2017elderly}. Schütz et al. extended this direction using in-home ambient sensors and self-training domain adaptation for visit detection, comparing one-class support vector machine (SVM) baselines with logistic regression and random forest classifiers \cite{schutz2021sensor}. Walsh et al. used ambient smart-home sensors, including motion, contact, electricity, water flow, and light-switch sensors, and applied linear discriminant analysis (LDA) to infer health-related metrics such as loneliness and anxiety \cite{walsh2014inferring}. These works show how passive home-monitoring data can be transformed into behavioral markers of social connectedness and well-being.


In parallel, sensor-based ML has been used for functional decline, frailty, activity monitoring, and rehabilitation-related outcomes. Fan et al. combined wearable gait sensors with comprehensive geriatric assessment data and evaluated random forest, decision tree, naïve Bayes, neural network, and stochastic gradient descent models for frailty classification in older adults \cite{fan2023digital}. Giggins et al. used smartwatch and inertial sensor data collected by community-dwelling older adults at home, and applied ML classifiers to discriminate frailty status \cite{giggins2025unsupervised}. Palermo et al. introduced the Technology Integrated Health Management (TIHM) dataset for remote dementia monitoring, including in-home sensor and health data that support ML-based detection of health and behavioral changes \cite{palermo2023tihm}. These studies demonstrate the use of wearable and ambient sensors for mobility, frailty, and functional-status monitoring.

Studies closer to lower-limb fracture recovery and post-discharge functional monitoring remain limited. North et al. used wearable ground reaction force sensors embedded in insoles, extracted step-level loading features, and applied logistic regression to predict fracture healing progression \cite{north2024predicting}. Within the Multimodal Artificial Intelligence-based Sensor platform for Older iNdividuals (MAISON) platform \cite{abedi2022maison}, Abedi et al. introduced a multimodal dataset from older adults recovering from lower-limb fractures, including smartphone, smartwatch, motion, sleep, mobility, heart rate, and clinical assessment data \cite{abedi2025multimodal}. Supervised ML and deep learning (DL) models were used mainly for technical validation and recovery-related prediction, rather than joint modeling of multiple clinical outcomes \cite{abedi2025multimodal}. A subsequent MAISON-based work linked the longitudinal sensor and clinical data with neighbourhood-level geospatial information to create GEOFRAIL, which included participant demographics; sensor-derived features; repeated frailty, physical function, and social isolation assessments; temporal location records; neighbourhood amenities; crime rates; and socioeconomic indicators \cite{abedi2026longitudinal}. Its results focused on dataset validation, consistency among geospatial, sensor-derived, and clinical measures, and baseline ML analyses (single output) rather than multi-output regression \cite{abedi2026longitudinal}. Khan et al. later used MAISON features, including acceleration, step count, ambient motion, global positioning system (GPS) location, heart rate, sleep, and clinical scores, with modality-wise multiview clustering and large language model (LLM)-based interpretation to explain recovery trajectories; however, the outputs were recovery clusters and narrative explanations rather than numerical joint predictions of social isolation and functional outcomes \cite{khan2025explaining}.

\subsection{Multi-Output and Multi-Task Learning in Health-Related Prediction}\label{sec:related-work-multi-Output}
Multi-output learning and multi-task learning (MTL) have been explored in several health-related prediction problems where multiple clinical outcomes share underlying structure. In Alzheimer’s disease research, Zhang and Shen proposed multi-modal multi-task learning to jointly predict regression and classification outcomes from magnetic resonance imaging, positron emission tomography (PET), cerebrospinal fluid, and clinical data \cite{zhang2012multi}. Zhou et al. formulated Alzheimer’s disease progression as a multi-task regression problem using temporal group Lasso regularization \cite{zhou2012modeling}. El-Sappagh et al. extended this direction using a multimodal MTL model with convolutional neural network and bidirectional long short-term memory components to jointly predict Alzheimer’s disease progression and cognitive scores from longitudinal multimodal data \cite{el2020multimodal}. These studies show that shared representations can be learned across related regression and classification targets in neuroimaging and cognitive progression.

Multi-task and multi-output learning have also been applied to electronic health record (EHR)-based prediction. Harutyunyan et al. introduced intensive care unit (ICU) benchmarks from Medical Information Mart for Intensive Care III (MIMIC-III), including mortality, decompensation, length of stay, and phenotype prediction, and evaluated linear and recurrent neural network models in single-task and multitask settings \cite{harutyunyan2019multitask}. Shickel et al. proposed multimodal transformers for jointly predicting multiple ICU outcomes \cite{shickel2021multi}, while Chan et al. developed a heterogeneous graph neural network (GNN) framework for simultaneous EHR tasks, including drug recommendation, length-of-stay prediction, mortality prediction, and readmission prediction \cite{chan2024multi}. Other studies used multi-output or multitask formulations for healthcare resource utilization \cite{cui2018prediction}, electronic phenotyping \cite{ding2019effectiveness}, and joint disease prediction from EHR data \cite{cui2024automated}.

MTL has also been applied to sensor-based prediction problems, particularly in wearable sensing and human activity recognition (HAR), although some of these works are not strictly health-related. Chen et al. proposed METIER, a deep MTL model using wearable sensors to jointly perform activity recognition and user recognition \cite{chen2020metier}; Nisar et al. used wearable sensor data with a CNN and long short-term memory based hierarchical MTL framework to jointly recognize state, behavioral, and composite activities of daily living \cite{nisar2023hierarchical}; and Duan et al. developed a multi-task DL model for sensor-based HAR that jointly performed activity segmentation and activity recognition from continuous sensor streams \cite{duan2023multitask}. More health-oriented sensor-based examples include Khan et al., who used gait analysis data and LSTM-based MTL to predict knee and ankle kinematic trajectories after botulinum toxin treatment in gait rehabilitation \cite{khan2022treatment}, and Saylam and {\.I}ncel, who used wearable-derived physical activity, sleep, and social-network-related features from the NetHealth dataset, comparing random forest, XGBoost, LSTM, and MTL models to predict depression, anxiety, and stress \cite{saylam2024multitask}.

\subsection{Research Gap and Contributions}\label{sec:related-work-gap}
The reviewed literature shows two related but largely separate research directions. First, sensor-based studies in older adults have used mobile, wearable, ambient, sleep, gait, and geospatial data to monitor or predict loneliness, social isolation, home visits, frailty, activity, fracture healing, and recovery trajectories \cite{austin2016smart,martinez2020automatic,goonawardene2017sensor,hu2017elderly,schutz2021sensor,walsh2014inferring,kang2025exploring,ji2026smile,prabhu2022sensor,qirtas2022loneliness,parraga2026sensor,fan2023digital,giggins2025unsupervised,north2024predicting,abedi2025multimodal,abedi2026longitudinal,khan2025explaining}. However, these studies usually define the learning problem around one clinical target at a time, such as loneliness status, social isolation risk, social interaction group, visit occurrence, frailty status, activity class, fracture healing status, or recovery cluster. Second, multi-output and MTL studies have shown the value of jointly modelling related targets, but they have primarily focused on neuroimaging, biomedical measurements, EHR data, sensor-based HAR, gait trajectories, or mental health prediction rather than post-discharge recovery in older adults after lower-limb fracture \cite{zhang2012multi,zhou2012modeling,el2020multimodal,harutyunyan2019multitask,shickel2021multi,chan2024multi,cui2018prediction,ding2019effectiveness,cui2024automated,chen2020metier,nisar2023hierarchical,duan2023multitask,khan2022treatment,saylam2024multitask}. Third, newer tabular deep-learning models have received limited evaluation for single-output and multi-output regression in community-based recovery monitoring.

Therefore, there remains a gap in applying multi-output regression to multimodal sensor-based prediction of social isolation and functional recovery in community-dwelling older adults. This setting is well suited for multi-output learning because the target outcomes reflect related dimensions of the same post-discharge recovery process. Social Isolation Scale (SIS) captures social participation and isolation risk, while Oxford Hip Score (OHS) and Oxford Knee Score (OKS) capture lower-limb pain and function \cite{cihi2024proms}. Timed Up and Go (TUG) reflects mobility, balance, walking ability, and fall risk, and Chair Stand captures lower-extremity strength and functional capacity \cite{beauchet2011timed,jones199930}. Prior evidence also supports the clinical linkage among these domains: social isolation and loneliness are associated with poorer physical performance, including sit-to-stand ability, balance, and walking speed; social isolation is associated with worse outcomes after hip fracture; and hip-fracture-related movement limitations can disrupt social life and contribute to isolation \cite{philip2020social,mandl2024effect,zare2024social}. These relationships suggest that the outcomes represent related dimensions of recovery and motivate an empirical comparison of independent and joint prediction approaches. Therefore, the present study  jointly predicts SIS, OHS, OKS, TUG, and Chair Stand from multimodal sensor-derived features and compares single- and multi-output regression across traditional ML and tabular DL models, following the broader rationale that related clinical targets can benefit from shared representations \cite{zhang2012multi,zhou2012modeling,el2020multimodal,harutyunyan2019multitask,shickel2021multi,chan2024multi,abedi2025multimodal,abedi2026longitudinal,khan2025explaining}.

\section{Methods}

\subsection{MAISON-LLF Dataset}
\label{sec:maison-llf}
The MAISON-LLF dataset \cite{abedi2025multimodal} contains longitudinal multimodal sensor and clinical assessment data collected from older adults recovering in the community after lower-limb fracture or hip replacement. Participants were recruited before hospital discharge, and data collection began within the first few days after discharge and continued for eight weeks. The MAISON platform \cite{abedi2022maison} integrated a Google Pixel Watch 2 smartwatch, Motorola Moto G54 smartphone, Proteus M5 in-home motion sensor, and Withings Sleep under-mattress sleep-tracking mat. The smartwatch collected acceleration, heart rate, step count, and GPS-based out-of-home mobility data; the motion sensor captured in-home movement events; and the sleep mat recorded sleep information. Clinical and functional assessments were collected every two weeks, including the SIS \cite{nicholson2020psychometric}, in which higher scores correspond to greater social interaction and lower social isolation; the OHS \cite{wylde2005oxford}, which assesses hip-related pain and function; the OKS \cite{whitehouse2005oxford}, which assesses knee-related pain and function; the TUG \cite{podsiadlo1991timed}, which measures functional mobility and balance; and the 30-second Chair Stand test \cite{chen2009normative}, which evaluates lower-extremity strength and functional capacity.
For each clinical assessment obtained every two weeks, the recorded outcome value was assigned to all daily sensor observations within the 14-day period ending on the assessment date, such that the 14 daily observations in each interval shared the same clinical outcome value.

The dataset was collected in the Greater Toronto Area, Canada, from participants recruited at Toronto Rehabilitation Institute, University Health Network. The study was approved by the University Health Network Research Ethics Board (study ID: 20-5113.10). Data collection was conducted between March 2022 and October 2025. The dataset includes daily sensor-derived features (refer to Table \ref{tab_features}) from 18 participants over 56 days, resulting in 1,008 days of multimodal sensor monitoring, together with repeated clinical questionnaire and physical performance assessments.

The study cohort had a mean age of 76.5 years (SD 8.9; range: 60--94), with 14 female participants (78\%), 14 participants living alone (78\%), and 15 participants identifying as Caucasian (83\%). The index procedure or fracture type included hip replacement in 8 participants (44\%), hip fracture in 4 (22\%), pelvis fracture in 3 (17\%), and femur fracture in 3 (17\%). Across all biweekly visits, the mean SIS score was 23.8 (SD 3.9; range: 15--30), OHS was 28.4 (SD 7.9; range: 17--47), OKS was 30.3 (SD 9.7; range: 15--48), TUG was 19.6 seconds (SD 12.6; range: 7--59), and Chair Stand performance was 9.8 repetitions (SD 3.4; range: 1--17).

\begin{table*}
\caption{The data modalities, feature names, and their description. $^1$Feature naming convention in the dataset: ‘data modality’ +‘-’+‘feature’.}
\label{tab_features}
\centering
\setlength{\tabcolsep}{4.5pt}
\begin{tabular}{|l|l|l|}
\hline
Data modality$^1$                   & Feature$^1$                                   & Description     \\ \hline
\multirow{7}{*}{acceleration-}  &  \makecell[l]{count,   entropy, kurtosis, mean, \\skew, std, and sum}      & \makecell[l]{The total count, entropy, kurtosis, mean, skewness, standard deviation, and sum of acceleration data \\in a day.}          \\ \cline{2-3}
                                & coefficient-of-variation                                                                      & \makecell[l]{Measures the relative variability of daily physical activity, indicating how consistently an individual \\moves throughout the day in relation to their average activity level.}          \\ \cline{2-3}
                                & minutes-with-data                                                                             & \makecell[l]{Represents the total number of minutes within a 24-hour period (from 12:00 AM to the following \\12:00 AM) during which acceleration data was recorded, providing insight into the smartwatch's \\active recording time, even if it was not necessarily worn on the wrist.}                           \\ \cline{2-3}
                                & hours-with-data                                                                               & \makecell[l]{Represents the total number of hours within a 24-hour period (from 12:00 AM to the following \\12:00 AM) during which acceleration data was recorded, providing insight into the smartwatch's \\active recording time, even if it was not necessarily worn on the wrist.}                             \\ \cline{2-3}
                                & \makecell[l]{movement-events-00to06,\\movement-events-06to12, \\movement-events-12to18, \\movement-events-18to24} & \makecell[l]{Represents the number of detected movement events between 12:00 AM and 6:00 AM, 6:00 AM  \\and 12:00 PM, 12:00 PM and 18:00 PM, 18:00 PM and 12:00 AM, based on acceleration values \\exceeding a dynamic threshold, and reflects physical activity or restlessness.}                                  \\ \cline{2-3}
                                & movement-events-24h                                                                           & \makecell[l]{Represents the total number of detected movement events over a full 24-hour period (from \\12:00 AM to the following 12:00 AM), based on acceleration values exceeding a dynamic \\threshold, and provides an overall measure of daily physical activity.}                                            \\ \cline{2-3}
                                & intradaily-variability                                                                        & \makecell[l]{Quantifies the fragmentation of daily activity patterns by measuring the frequency and extent of \\transitions between active and inactive periods within a 24-hour   day; higher values indicate more \\erratic or irregular activity rhythms.}                                                          \\ \hline
\multirow{2}{*}{heartrate-}     & count,   max, mean, min, and std                                                              & The count, maximum, mean, minimum, and standard deviation of heartrate in a day.       \\ \cline{2-3}
                                & hours-with-data                                                                               & \makecell[l]{Represents the number of distinct hours within a 24-hour period (from 12:00 AM to the following \\12:00 AM) during which at least one heart rate measurement was recorded, reflecting periods of \\active heart rate monitoring and providing an estimate of smartwatch wear time (user compliance).} \\ \hline
\multirow{5}{*}{motion (step)-} & count                                                                                         & The total count of motions (steps) per day.               \\ \cline{2-3}
                                & max                                                                                           & The maximum number of motions (steps) in hours of a day.     \\ \cline{2-3}
                                & max-timestamp                                                                                 & The timestamp (hour of day) in which there has been maximum number of motions   (steps) in a day.         \\ \cline{2-3}
                                & mean                                                                                          & The average number of motions (steps) in hours of a day with at least one motion   (step).               \\ \cline{2-3}
                                & ratio                                                                                         & \makecell[l]{The ratio of the number of hours with at least a motion (step) to the number of hours without any \\motion (step) in a day. }        \\ \hline
\multirow{3}{*}{position-}      & count                                                                                         & The   total count of position data in a day.           \\ \cline{2-3}
                                & duration                                                                                      & The duration (in hours) of being outside the home in a day.           \\ \cline{2-3}
                                & distance-travelled                                                                            & The total distance (in kilometers) travelled outside the home in a day.                   \\ \hline
\multirow{4}{*}{sleep-}         & \makecell[l]{deep,   light, rem, snoring, and \\total}                                                        & The duration (in hours) of deep, light, rapid-eye-movement, snoring, and total sleep.       \\ \cline{2-3}
                                & \makecell[l]{duration-to-sleep and \\duration-to-wakeup}                                                    & The duration (in hours) to sleep and to wake up.         \\ \cline{2-3}
                                & \makecell[l]{heartrate-max,   heartrate-mean, \\and heartrate-min}                                            & The maximum, mean, and minimum heartrate during sleep.        \\ \cline{2-3}
                                & wakeup-count                                                                                  & The count of wakeups during sleep time.     \\ \hline                                                    
\end{tabular}
\end{table*}

\subsection{Experimental setup}
To comprehensively evaluate the predictive performance for social isolation and functional decline, the analysis in Section \ref{sec_results} used a range of traditional machine learning (ML) and state-of-the-art tabular deep learning (DL) regressors. The traditional ML regressors included Linear Regression (LR), ExtraTrees Regressor (ET), Random Forest Regressor (RF), K-Nearest-Neighbors regressor (KNN), and Support Vector Regressor (SVR), with their implementations from the corresponding sklearn library. The DL regressors included NODE \cite{popovneural}, FT-Transformer \cite{gorishniy2021revisiting}, TabPFN \cite{hollmanntabpfn}, and TabNet \cite{arik2021tabnet}. The core objective of the analysis was to investigate the performance of multi-output regression compared to single-output regression when predicting the target clinical outcomes. The target outcomes were the Social Isolation Scale (SIS), Oxford Hip Score (OHS), Oxford Knee Score (OKS), Timed Up and Go (TUG), and Chair Stand. While single-output regression isolated these variables and predicted them individually, multi-output regression simultaneously predicted all five variables, allowing the models to leverage the correlations for enhanced predictive accuracy. The sklearn implementations of the ML models LR, ET, RF, and KNN have in-built support for multi-output regression, while SVR used sklearn RegressorChain wrapper to run multi-output regression. All the ML models used sklearn MultiOutputRegressor wrapper to implement single-output regression. As the range of values varied widely across the target variables, a min-max scaling transformation was done before training the models to bring all the target variables in the same range. The target variable and predicted values were then transformed back to the original range before calculating the evaluation metrics. The evaluation metrics were Mean Squared Error (MSE) and Mean Absolute Error (MAE) between the predicted values and the target variable values. The ML regressors were trained with the default parameters. 
As the physical health and behavioural routines vary across different individuals and time periods, the models were evaluated using two distinct and rigorous validation setups.
\begin{itemize}
    \item Leave-one-person-out cross-validation: The dataset was divided by participant folds, and models were iteratively trained on all but one participant, testing on the held-out unseen individual. This tested the model's ability to safely generalize to completely new older adults without prior exposure to their specific baseline behaviours.
    \item Leave-one-week-out cross-validation: This strategy holds out a specific week of data across the entire cohort. This temporal validation tested the models' robustness against week-to-week behavioural shifts, demonstrating that the model does not memorize time-specific routines.
\end{itemize}

\section{Results and Discussion} \label{sec_results}

\subsection{Correlations among clinical outcomes}
\begin{figure}
\centering
\includegraphics[width=\columnwidth]{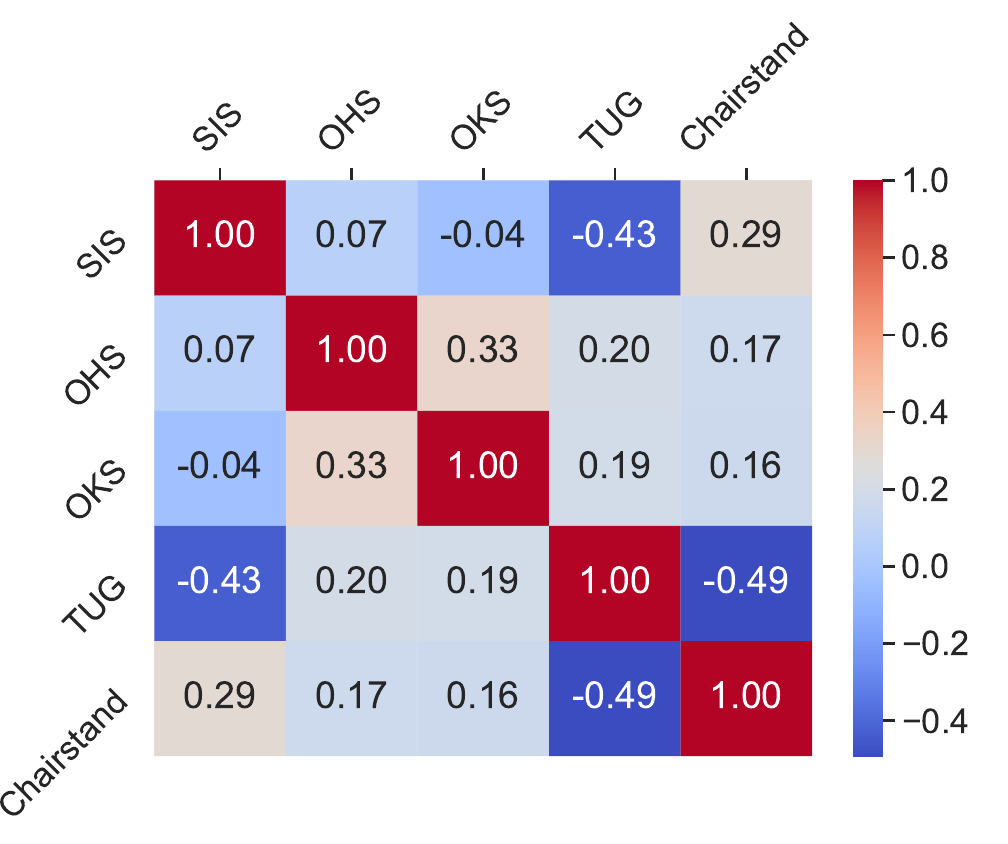}
\caption{Correlation matrix of the clinical outcomes for the overall dataset.}
\label{fig_corrOver}
\end{figure}

In previous sections, we mentioned the structural similarity between various functional mobility and social isolation scales as they represent different pathways around the recovery of this population. To verify this in the MAISON-LLF dataset, we performed a correlation analysis.  
Figure \ref{fig_corrOver} illustrates the correlation matrix for the overall dataset, revealing significant coupled relationships between functional mobility and social isolation. Most notably, the TUG test exhibits a strong negative correlation with both the Chair Stand test (-0.49) and SIS (-0.43). Furthermore, SIS demonstrates a positive correlation with Chair Stand performance (0.29). On the joint health side, the OHS and OKS share a clear positive correlation (0.33). These overall metrics indicate that functional decline and social isolation may not occur independently; rather, they represent a high correlation of physiological and behavioural markers. 

\begin{figure*}
\centering
\includegraphics[width=\textwidth]{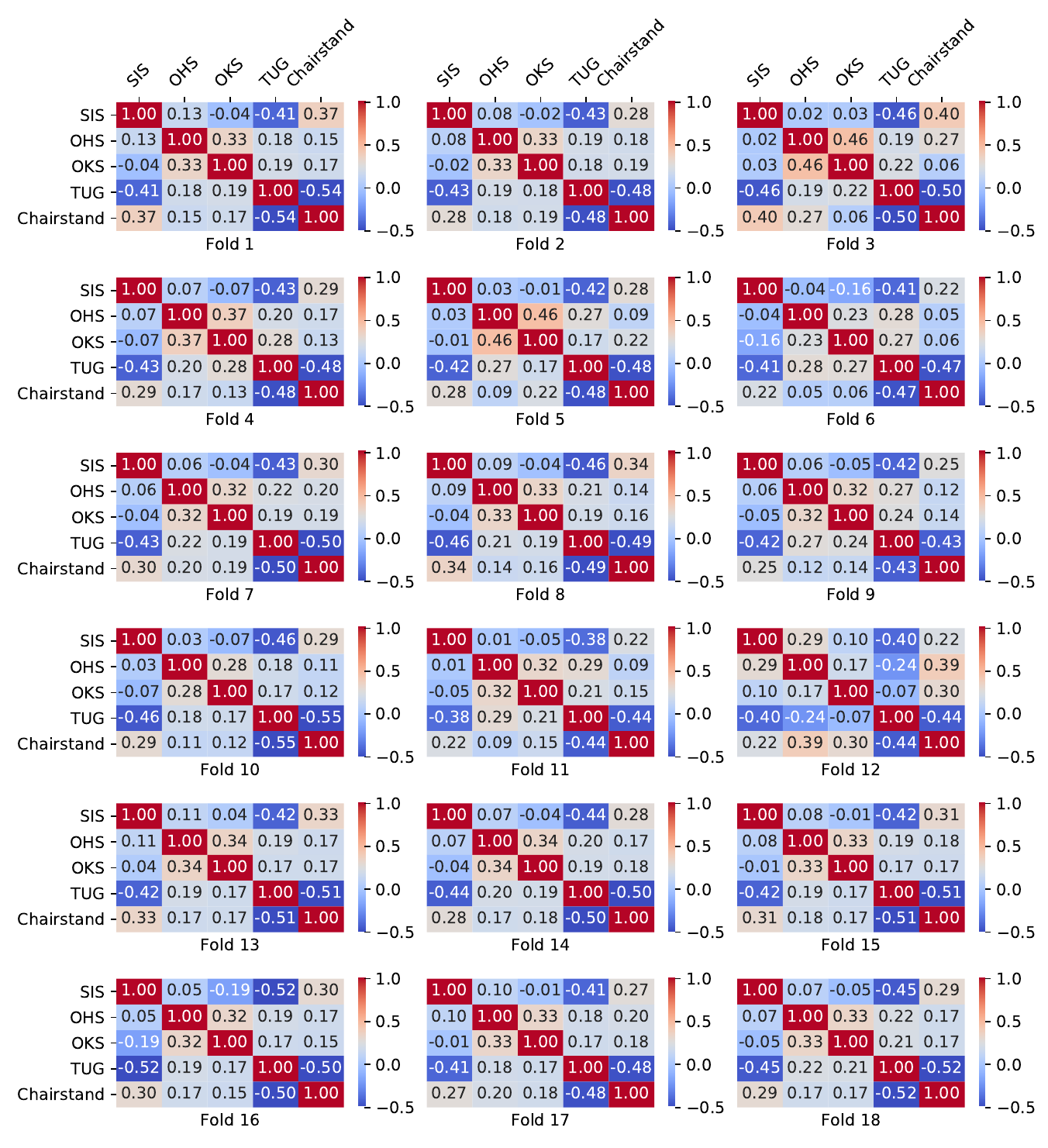}
\caption{Correlation matrix of the outcomes for each individual fold as part of the leave-one-participant-out cross validation.}
\label{fig_corrFolds}
\end{figure*}

Further, to ensure that these correlations are intrinsic to the studied cohort and not merely an artifact of a few outliers, we extracted individual correlation matrices for all 18 participant folds as part of the leave-one-participant-out cross-validation methodology. As demonstrated in Figure \ref{fig_corrFolds}, these relationships remain remarkably stable across individual, unseen participants. For instance, the strong negative correlation between TUG and Chair Stand persists consistently across the population, ranging from -0.43 in Fold 9 to as strong as -0.55 in Fold 10. Similarly, the negative correlation between SIS and TUG remains highly robust across individual folds, reaching up to -0.52 in Fold 16. The interconnected nature of joint health also holds firm, with the positive correlation between OHS and OKS peaking at 0.46 in Folds 3 and 5.

The stability of these correlations across the 18 independent participant folds validates our assertion and underpins the necessity for a multi-output predictive approach. The single-output regression models artificially isolate these target variables, mathematically discarding the vital contextual information that a decline in functional mobility (e.g., TUG) is statistically coupled with increased social isolation (SIS). By simultaneously predicting all five variables, the proposed multi-output ML and DL methods are forced to explicitly map and exploit these shared representations, yielding a more robust, context-aware, and clinically plausible predictive outcomes.

\begin{table*}[h]
\caption{Results of multi-output and single-output regression for leave-one-person-out cross-validation in terms of (a) MSE and (b) MAE. Unless the errors are equivalent, the best values for each outcome are highlighted in bold. The blue shaded cells signify the better values between single and multi-output regression.}
\label{tab_LOPO}
\centering
\setlength{\tabcolsep}{4pt}
\subfloat[MSE]{%
\begin{tabular}{|l|l|l|l|l|l|l|l|l|l|l|l|l|}
\hline
\multicolumn{1}{|l}{\multirow{2}{*}{}}           & \multicolumn{6}{|c}{Multi-output Regression}                                                                        & \multicolumn{6}{|c|}{Single-output Regression}                                                       \\ \cline{2-13}
\multicolumn{1}{|l|}{}                            & SIS            & OHS            & OKS             & TUG             & Chair Stand     & Average & SIS            & OHS             & OKS            & TUG             & Chair Stand & Average         \\ \hline
\multicolumn{13}{|c|}{ML Regressors}                                                                                                                                                                                                                          \\ \hline
LR        & 23.32          & 226.92         & 130.93          & 473.98          & 31.28          & 177.28                      & 23.32          & 226.92          & 130.93         & 473.98          & 31.28      & 177.28          \\ \hline
ET     & 17.38          & 91.06 & 100.12          & 309.97          & 24.72 & 108.65             & 15.95 & 103.73          & 99.16 & 307.21 & 27.52      & 110.71          \\ \hline
RF  & 17.61          & 86.14 & 105.40 & 303.24 & 24.96 & 107.47             & 16.74 & 107.08          & 109.56         & 350.50          & 27.99      & 122.37          \\ \hline
KNN     & 18.75          & 119.46         & 134.32          & 438.73          & 34.17          & 149.09                      & 18.75          & 119.46          & 134.32         & 438.73          & 34.17      & 149.09          \\ \hline
SVR & 19.49          & 111.72         & 115.02 & 365.04          & 28.02 & 127.86                      & 19.49          & 108.92 & 123.18         & 351.66 & 29.73      & 126.60 \\ \hline
\multicolumn{13}{|c|}{DL Regressors}                                                                                                                                                                                                                             \\ \hline
NODE                     & \cellcolor{blue!25} \textbf{4.60}  & \cellcolor{blue!25} \textbf{9.32}  & \cellcolor{blue!25} \textbf{6.01}   & \textbf{71.46}  & \cellcolor{blue!25} \textbf{3.96}  & \cellcolor{blue!25} \textbf{19.07}              & 12.79          & \textbf{12.80}           & 30.53          & 117.87          & \textbf{6.80}       & 36.16           \\ \hline
FT-Transformer           & 5.04  & 26.12          & 24.84           & 92.96  & 7.74  & 31.34              & 7.50           & 23.41  & \textbf{19.95} & 131.87          & 8.89       & 38.33           \\ \hline
TabPFN                   & 5.03  & 12.26 & 13.57  & 74.04           & 5.06  & 21.99              & \textbf{7.43}           & 16.44           & 30.19          & \cellcolor{blue!25} \textbf{58.86}  & 9.67       & \textbf{24.52}           \\ \hline
TabNet                   & 15.08 & 73.51 & 122.40 & 318.20 & 28.20 & 111.48             & 19.02          & 85.78           & 255.87         & 326.50          & 52.20      & 147.87          \\ \hline
\end{tabular}
}
\quad 
\subfloat[MAE]{%
\begin{tabular}{|l|l|l|l|l|l|l|l|l|l|l|l|l|}
\hline
\multicolumn{1}{|l}{\multirow{2}{*}{}}           & \multicolumn{6}{|c}{Multi-output Regression}                                                                        & \multicolumn{6}{|c|}{Single-output Regression}                                                       \\ \cline{2-13}
\multicolumn{1}{|l|}{}                            & SIS            & OHS            & OKS             & TUG             & Chair Stand     & Average & SIS            & OHS             & OKS            & TUG             & Chair Stand & Average         \\ \hline
\multicolumn{13}{|c|}{ML Regressors}                                                                                                                                                                                                                          \\ \hline
LR        & 3.92           & 9.89           & 9.49            & 16.29           & 4.68           & 8.85                        & 3.92           & 9.89            & 9.49           & 16.29           & 4.68       & 8.85            \\ \hline
ET     & 3.42           & 7.60  & 8.31            & 12.10           & 4.25  & 7.14               & 3.31  & 8.10            & 8.18  & 11.80  & 4.51       & 7.18            \\ \hline
RF  & 3.35  & 7.24  & 8.28   & 11.00  & 4.20  & 6.82               & 3.42           & 8.38            & 8.47           & 12.72           & 4.51       & 7.50            \\ \hline
KNN     & 3.60           & 8.56           & 9.59            & 14.83           & 4.82           & 8.28                        & 3.60           & 8.56            & 9.59           & 14.83           & 4.82       & 8.28            \\ \hline
SVR & 3.70           & 8.33           & 9.08   & 14.04           & 4.57  & 7.94                        & 3.70           & 8.26   & 9.30           & 13.39  & 4.71       & 7.87   \\ \hline
\multicolumn{13}{|c|}{DL Regressors}                                                                                                                                                                                                                             \\ \hline
NODE                     & \cellcolor{blue!25} \textbf{1.06}  & \cellcolor{blue!25} \textbf{1.77}  & \cellcolor{blue!25} \textbf{1.52}   & 4.44   & \cellcolor{blue!25} \textbf{1.02}  & \cellcolor{blue!25} \textbf{1.96}               & 2.83           & \textbf{2.53}            & 3.76           & 6.25            & \textbf{1.89}       & 3.45            \\ \hline
FT-Transformer           & 1.55  & 3.43  & 3.48            & 5.84   & 1.87  & 3.23               & \textbf{1.95}           & 3.52            & \textbf{3.35}  & 6.58            & 2.22       & 3.53            \\ \hline
TabPFN                   & 1.64  & 2.67  & 2.65   & \textbf{4.32}            & 1.49  & 2.56               & 2.10           & 3.04            & 4.14           & \cellcolor{blue!25} \textbf{3.85}   & 2.39       & \textbf{3.11}            \\ \hline
TabNet                   & 3.16  & 6.92  & 9.29   & 13.02           & 4.37  & 7.35               & 3.73           & 7.19            & 11.63          & 12.56  & 5.25       & 8.07    \\ \hline
\end{tabular}
}
\end{table*}

\begin{table}[h]
\centering
\caption{Percentage improvement of multi-output over single-output regressors for leave-one-person-out cross-validation. The best values for each outcome across all the regressors are highlighted in bold.}
\label{tab_percent_LOPO}
\setlength{\tabcolsep}{4pt}
\subfloat[MSE]{%
\begin{tabular}{|l|l|l|l|l|l|l|}
\hline
\multicolumn{1}{|l|}{}                            & SIS   & OHS    & OKS    & TUG    & Chair Stand & Average \\ \hline
\multicolumn{7}{|c|}{ML Regressors}                  \\ \hline
LR        & 0     & 0      & 0      & 0      & 0          & 0               \\ \hline
ET     & -8.97 & 12.22  & -0.97  & -0.9   & 10.17      & 2.31            \\ \hline
RF  & -5.2  & 19.55  & 3.8    & 13.48  & 10.8       & 8.49            \\ \hline
KNN     & 0     & 0      & 0      & 0      & 0          & 0               \\ \hline
SVR & 0     & -2.57  & 6.63   & -3.81  & 5.77       & 1.2             \\ \hline
\multicolumn{7}{|c|}{DL Regressors}                     \\ \hline
NODE                     & \textbf{64.06} & \textbf{27.18}  & \textbf{80.3}   & \textbf{39.38}  & 41.8       & \textbf{50.54}           \\ \hline
FT-Transformer           & 32.9  & -11.58 & -24.52 & 29.5   & 12.96      & 7.85            \\ \hline
TabPFN                   & 32.32 & 25.45  & 55.03  & -25.79 & \textbf{47.72}      & 26.95           \\ \hline
TabNet                   & 20.74 & 14.31  & 52.16  & 2.54   & 45.98      & 27.15           \\ \hline
\end{tabular}
}
\quad 
\subfloat[MAE]{%
\begin{tabular}{|l|l|l|l|l|l|l|}
\hline
\multicolumn{1}{|l|}{}                            & SIS    & OHS    & OKS   & TUG   & Chair Stand & Average \\ \hline
\multicolumn{7}{|c|}{ML Regressors}                  \\ \hline
LR        				 & 0     & 0      & 0      & 0      & 0          & 0                \\ \hline
ET     					 & -3.35 & 6.2    & -1.68  & -2.59  & 5.92       & 0.9              \\ \hline
RF   					 & 1.9   & 13.53  & 2.2    & 13.57  & 6.87       & 7.61             \\ \hline
KNN     				 & 0     & 0      & 0      & 0      & 0          & 0                \\ \hline
SVR 					 & 0     & -0.79  & 2.41   & -4.82  & 2.91       & -0.06            \\ \hline
\multicolumn{7}{|c|}{DL Regressors}                     \\ \hline
NODE                     & \textbf{62.76} & \textbf{30.15}  & \textbf{59.5}   & \textbf{29.06}  & \textbf{46.33}      & \textbf{45.56}            \\ \hline
FT-Transformer           & 20.56 & 2.75   & -3.73  & 11.21  & 15.86      & 9.33             \\ \hline
TabPFN                   & 21.83 & 12.29  & 35.82  & -12.12 & 37.64      & 19.09            \\ \hline
TabNet                   & 15.3  & 3.85   & 20.12  & -3.65  & 16.72      & 10.47           \\ \hline
\end{tabular}
}
\end{table}

\begin{table}[h]
\caption{Percentage improvement of multi-output regressors with respect to LR baseline for leave-one-person-out cross-validation. The best values for each outcome across all the regressors are highlighted in bold.}
\label{tab_percent_overLR_LOPO}
\centering
\setlength{\tabcolsep}{4pt}
\subfloat[MSE]{%
\begin{tabular}{|l|l|l|l|l|l|l|}
\hline
\multicolumn{1}{|l|}{}                            & SIS   & OHS   & OKS   & TUG   & Chair Stand & Average \\ \hline
\multicolumn{7}{|c|}{ML Regressors}              \\ \hline
LR        				 & 0     & 0     & 0     & 0     & 0          & 0             \\ \hline
ET     					 & 25.47 & 59.87 & 23.53 & 34.6  & 20.96      & 32.89         \\ \hline
RF  					 & 24.49 & 62.04 & 19.5  & 36.02 & 20.18      & 32.45         \\ \hline
KNN     				 & 19.58 & 47.35 & -2.59 & 7.44  & -9.26      & 12.5          \\ \hline
SVR 					 & 16.4  & 50.77 & 12.15 & 22.98 & 10.42      & 22.54         \\ \hline
\multicolumn{7}{|c|}{DL Regressors}                 \\ \hline
NODE                     & \textbf{80.29} & \textbf{95.89} & \textbf{95.41} & \textbf{84.92} & \textbf{87.35}      & \textbf{88.77}         \\ \hline
FT-Transformer           & 78.41 & 88.49 & 81.03 & 80.39 & 75.25      & 80.71         \\ \hline
TabPFN                   & 78.43 & 94.6  & 89.63 & 84.38 & 83.83      & 86.17         \\ \hline
TabNet                   & 35.34 & 67.61 & 6.52  & 32.87 & 9.84       & 30.44         \\ \hline
\end{tabular}
}
\quad 
\subfloat[MAE]{%
\begin{tabular}{|l|l|l|l|l|l|l|}
\hline
\multicolumn{1}{|l|}{}                            & SIS   & OHS   & OKS   & TUG   & Chair Stand & Average \\ \hline
\multicolumn{7}{|c|}{ML Regressors}              \\ \hline
LR        				 & 0     & 0     & 0     & 0     & 0          & 0              \\ \hline
ET     					 & 12.59 & 23.21 & 12.4  & 25.67 & 9.37       & 16.65          \\ \hline
RF  					 & 14.35 & 26.77 & 12.71 & 32.48 & 10.27      & 19.32          \\ \hline
KNN     				 & 8.07  & 13.49 & -1.04 & 8.93  & -2.92      & 5.31           \\ \hline
SVR 					 & 5.52  & 15.79 & 4.31  & 13.8  & 2.46       & 8.38           \\ \hline
\multicolumn{7}{|c|}{DL Regressors}                 \\ \hline
NODE                     & \textbf{73.06} & \textbf{82.13} & \textbf{83.96} & 72.76 & \textbf{78.31}      & \textbf{78.04}          \\ \hline
FT-Transformer           & 60.44 & 65.35 & 63.34 & 64.14 & 60.12      & 62.68          \\ \hline
TabPFN                   & 58.02 & 73.01 & 72.03 & \textbf{73.48} & 68.17      & 68.94          \\ \hline
TabNet                   & 19.41 & 30.08 & 2.11  & 20.05 & 6.75       & 15.68          \\ \hline
\end{tabular}
}
\end{table}

\subsection{Leave-one-person-out cross-validation}
We evaluate the predictive capabilities of the models on unseen older adults by analyzing the MSE and MAE under the leave-one-person-out cross-validation methodology based on error magnitudes in Table \ref{tab_LOPO} and the relative percentage improvements in Table \ref{tab_percent_LOPO}. 
The results demonstrate that compared to single-output, multi-output regression yields substantial performance gains, particularly when leveraging DL architectures. 
Wilcoxon signed-rank test was performed to evaluate the statistical significance of the performance differences between multi-output and single-output regression. The performance improvement of multi-output over single-output was found to be statistically significant (p-value = $0.006$).

Under the leave-one-person-out cross-validation in Table \ref{tab_LOPO}, traditional ML baselines struggled to generalize to unseen participants; for instance, both the single-output and multi-output LR models yielded a high average MSE of 177.28 and an average MAE of 8.85. In contrast, the tabular DL models exhibited an ability to leverage the correlated functional decline and social isolation outcomes. The NODE regressor proved to be the most robust model. In single-output regression, NODE achieved an average MSE of 36.16; however, when tasked with predicting all five variables simultaneously (i.e. multi-output), its average MSE reduced to 19.07 (Table \ref{tab_LOPO}). This multi-output superiority is consistent across individual clinical variables. For example, when predicting the OKS, the multi-output NODE model achieved an MSE of 6.01, compared to 30.53 in the single-output model. Similarly, the prediction error for SIS dropped from an MSE of 12.79 in single-output to 4.60 in multi-output mode, validating that the functional decline correlation can be leveraged for predicting social isolation. The TUG score, which represents a functional mobility marker, also saw its prediction MSE reduced from 117.87 to 71.46 when modelled alongside the other variables.
These gains are further corroborated by the MAE metrics (Table \ref{tab_LOPO} (b)), where the multi-output NODE model achieved a low average MAE of 1.96, compared to 3.45 in the single-output setup. Other DL architectures mirrored this pattern; TabPFN and TabNet achieved average MSE of 2.56 and 7.35, respectively, for multi-output, compared to 3.11 and 8.07 for single output.
The analysis shows that while traditional ML models lack the representational capacity to effectively capture multi-output correlations (often resulting in marginal percentage improvements), DL models successfully model these shared dependencies. By leveraging a multi-output DL approach, we can generate accurate and context-aware multiple clinical score predictions for new, unseen participants.

To quantify the precise performance gains achieved by switching from single-output to multi-output prediction, we evaluated the percentage improvements across all regressors in Table \ref{tab_percent_LOPO}.
The percentage improvements reveal that traditional ML models generally lack the representational capacity to effectively capture and exploit multi-output correlations. While tree-based models, e.g., the RF, performed slightly better, they only achieved a modest 8.49\% average improvement in MSE in Table \ref{tab_percent_LOPO} when utilizing the multi-output approach, and even demonstrated a performance degradation (-5.2\%) when predicting the Social Isolation Scale (SIS). 
Conversely, the advanced DL architectures thrived mathematically on these shared dependencies. The NODE architecture secured a remarkable overall average MSE improvement of 50.54\% and an average MAE improvement of 45.56\% when switching to the multi-output paradigm. Examining specific clinical variables, NODE achieved a notable 80.3\% MSE performance improvement for OKS, a 64.06\% improvement for SIS, and a 39.38\% improvement for TUG. This validates that functional decline correlations can be leveraged for predicting social isolation.
Other tabular DL architectures mirrored this success; TabPFN and TabNet achieved average MSE improvements of 26.95\% and 27.15\%, respectively, when switching to multi-output regression. This percentage improvement analysis demonstrates that by leveraging a multi-output DL approach, we can successfully map shared clinical dependencies to generate accurate, context-aware predictions for new, unseen participants.

To further investigate the benefits of advanced models, we evaluated the percentage improvement of all multi-output regressors against the standard multi-output LR baseline model under the leave-one-person-out cross-validation in Table \ref{tab_percent_overLR_LOPO}. By setting the LR performance as the baseline, we can directly quantify the capacity of more complex algorithms to capture the non-linear physiological and behavioural relationships inherent in the multi-output predictive approach. Traditional ML models offered moderate, albeit notable, improvements over the linear baseline. Tree-based ensemble methods performed better, with the ET and RF regressors achieving average MSE improvements of 32.89\% and 32.45\%, respectively. Other ML models like the SVR and KNN yielded lower average MSE improvements of 22.54\% and 12.5\%.
However, the most significant finding is the notable performance of tabular DL models. The DL models outpaced both the LR baseline and the ML models, demonstrating an intrinsic capability to decode complex, correlated clinical variables. The multi-output NODE model achieved a 88.77\% average improvement in MSE across all five variables. Further, NODE's performance gains were high for the joint health metrics, exhibiting a 95.89\% improvement in OHS prediction and a 95.41\% improvement for the OKS relative to the LR baseline.
Other transformer-based DL models closely mirrored this predictive pattern. TabPFN and FT-Transformer delivered average MSE improvements of 86.17\% and 80.71\%, respectively. TabPFN also showed improvement on specific clinical markers, such as a 94.6\% MSE improvement for the OHS. TabNet was the only DL model that performed moderately, trailing the others with an average MSE improvement of 30.44\%.

These significant error reductions were corroborated by the MAE improvements (Table \ref{tab_percent_overLR_LOPO} (b)). The NODE regressor secured an average MAE improvement of 78.04\%, followed by TabPFN (68.94\%) and FT-Transformer (62.68\%). In contrast, the highest-performing traditional ML models, RF and ET, managed average MAE improvements of 19.32\% and 16.65\%.
This performance gap confirms that simple linear relationships may not be sufficient for multi-output prediction in this clinical context. Instead, the results show that advanced DL models, including NODE, TabPFN, and FT-Transformer are essential to leverage the predictive capacity of multi-output regression when evaluating complex markers of social isolation and functional decline.

\begin{table*}
\caption{Results of multi-output and single-output regression for leave-one-week-out cross-validation in terms of (a) MSE and (b) MAE. Unless the errors are equivalent, the best values for each outcome are highlighted in bold. The blue shaded cells signify the better values between single and multi-output regression.}
\label{tab_LOWO}
\centering
\setlength{\tabcolsep}{4pt}
\subfloat[MSE]{%
\begin{tabular}{|l|l|l|l|l|l|l|l|l|l|l|l|l|}
\hline
\multicolumn{1}{|l|}{\multirow{2}{*}{}}           & \multicolumn{6}{c|}{Multi-output Regression}                                                                       & \multicolumn{6}{c|}{Single-output Regression}                                                        \\ \cline{2-13}
\multicolumn{1}{|l|}{}                            & SIS            & OHS            & OKS            & TUG             & Chair Stand     & Average & SIS           & OHS            & OKS            & TUG             & Chair Stand     & Average        \\ \hline
\multicolumn{13}{|c|}{ML Regressors}       \\ \hline
LR        & 8.93           & 48.21          & 57.58          & 200.68          & 17.22          & 66.52                       & 8.93          & 48.21          & 57.58          & 200.68          & 17.22          & 66.52          \\ \hline
ET     & 4.87  & 19.33 & 21.29          & 91.33  & 7.09  & 28.79              & 5.28          & 20.31          & 21.07 & 102.24          & 8.05           & 31.39          \\ \hline
RF  & 5.40           & 18.43 & 22.58 & 98.74  & 7.82  & 30.59              & 5.23 & 25.39          & 23.05          & 108.97          & 9.93           & 34.51          \\ \hline
KNN     & 5.74           & 29.22          & 31.08          & 131.02          & 13.11          & 42.03                       & 5.74          & 29.22          & 31.08          & 131.02          & 13.11          & 42.03          \\ \hline
SVR & 6.56           & 31.51          & 37.58 & 145.24          & 12.08          & 46.59                       & 6.56          & 31.27 & 38.32          & 143.92 & 11.98 & 46.41 \\ \hline
\multicolumn{13}{|c|}{DL Regressors}                                                                                                                                                                                                                             \\ \hline
NODE                     & \cellcolor{blue!25} \textbf{0.38}  & \cellcolor{blue!25} \textbf{1.53}  & \cellcolor{blue!25} \textbf{1.81}  & \cellcolor{blue!25} \textbf{7.49}   & \cellcolor{blue!25} \textbf{0.67}  & \cellcolor{blue!25} \textbf{2.38}               & \textbf{0.64}          & 5.92           & \textbf{2.66}           & \textbf{13.82}           & \textbf{0.97}           & \textbf{4.80}           \\ \hline
FT-Transformer           & 1.11 & 5.58  & 3.90  & 18.11  & 1.54  & 6.05               & 1.94          & 67.70          & 9.37           & 46.79           & 4.23           & 26.01          \\ \hline
TabPFN                   & 1.95           & 6.42           & 5.69  & 40.88           & 1.87  & 11.36              & 1.70 & \textbf{3.56}  & 17.88          & 38.78  & 3.48           & 13.08          \\ \hline
TabNet                   & 10.30 & 45.08 & 56.13 & 196.00 & 16.26 & 64.75              & 22.25         & 92.40          & 95.53          & 507.70          & 38.44          & 151.26         \\ \hline
\end{tabular}
}
\quad 
\subfloat[MAE]{%
\begin{tabular}{|l|l|l|l|l|l|l|l|l|l|l|l|l|}
\hline
\multicolumn{1}{|l|}{\multirow{2}{*}{}}           & \multicolumn{6}{c|}{Multi-output Regression}                                                                       & \multicolumn{6}{c|}{Single-output Regression}                                                        \\ \cline{2-13}
\multicolumn{1}{|l|}{}                            & SIS            & OHS            & OKS            & TUG             & Chair Stand     & Average & SIS           & OHS            & OKS            & TUG             & Chair Stand     & Average        \\ \hline
\multicolumn{13}{|c|}{ML Regressors}       \\ \hline
LR        & 2.35           & 5.48           & 6.09           & 9.47   & 3.30           & 5.34                        & 2.35          & 5.48           & 6.09           & 9.47            & 3.30           & 5.34           \\ \hline
ET     & 1.61  & 3.12  & 3.43           & 4.94   & 1.91  & 3.00               & 1.67          & 3.24           & 3.39  & 5.22            & 2.03           & 3.11           \\ \hline
RF  & 1.70  & 3.10  & 3.39  & 4.98   & 1.96  & 3.02               & 1.70          & 3.59           & 3.45           & 5.21            & 2.19           & 3.23           \\ \hline
KNN     & 1.58           & 3.61           & 3.77           & 5.41            & 2.39           & 3.35                        & 1.58          & 3.61           & 3.77           & 5.41            & 2.39           & 3.35           \\ \hline
SVR & 1.99           & 4.27           & 4.73  & 8.11            & 2.65           & 4.35                        & 1.99          & 4.26  & 4.75           & 7.77   & 2.63  & 4.28  \\ \hline
\multicolumn{13}{|c|}{DL Regressors}                                                                                                                                                                                                                             \\ \hline
NODE                     & \cellcolor{blue!25} \textbf{0.24}  & \cellcolor{blue!25} \textbf{0.47}  & \cellcolor{blue!25} \textbf{0.51}  & \cellcolor{blue!25} \textbf{0.94}   & \cellcolor{blue!25} \textbf{0.31}  & \cellcolor{blue!25} \textbf{0.50}               & \textbf{0.34}          & 1.42           & \textbf{0.67}           & \textbf{1.53}            & \textbf{0.46}           & \textbf{0.88}           \\ \hline
FT-Transformer           & 0.52  & 1.05  & 0.97  & 1.92   & 0.61  & 1.01               & 0.89          & 6.20           & 1.83           & 3.53            & 1.60           & 2.81           \\ \hline
TabPFN                   & 0.96           & 1.73           & 1.69  & 3.80            & 0.96  & 1.83               & 0.92 & \textbf{1.03}  & 3.16           & 3.50   & 1.24           & 1.97           \\ \hline
TabNet                   & 2.56  & 5.14  & 5.95  & 9.38   & 3.18  & 5.24               & 3.96          & 7.75           & 8.13           & 16.45           & 5.01           & 8.26    \\ \hline
\end{tabular}
}
\end{table*}

\begin{table}[h]
\centering
\caption{Percentage improvement of multi-output over single-output regressors for leave-one-week-out cross-validation. The best values for each outcome across all the regressors are highlighted in bold.}
\label{tab_percent_LOWO}
\setlength{\tabcolsep}{4pt}
\subfloat[MSE]{%
\begin{tabular}{|l|l|l|l|l|l|l|}
\hline
\multicolumn{1}{|l|}{}                            & SIS    & OHS    & OKS   & TUG   & Chair Stand & Average \\ \hline
\multicolumn{7}{|c|}{ML Regressors}                  \\ \hline
LR        				 & 0      & 0      & 0     & 0     & 0          & 0       \\ \hline
ET     					 & 7.83   & 4.79   & -1.03 & 10.67 & 11.86      & 6.82    \\ \hline
RF  					 & -3.23  & 27.41  & 2.03  & 9.39  & 21.23      & 11.37   \\ \hline
KNN     				 & 0      & 0      & 0     & 0     & 0          & 0       \\ \hline
SVR 					 & 0      & -0.78  & 1.92  & -0.92 & -0.78      & -0.11   \\ \hline
\multicolumn{7}{|c|}{DL Regressors}                     \\ \hline
NODE                     & 40.84  & 74.17  & 31.83 & 45.83 & 30.97      & 44.73   \\ \hline
FT-Transformer           & 42.55  & \textbf{91.76}  & 58.39 & 61.3  & \textbf{63.53}      & \textbf{63.51}   \\ \hline
TabPFN                   & -14.44 & -80.43 & \textbf{68.17} & -5.42 & 46.19      & 2.81    \\ \hline
TabNet                   & \textbf{53.69}  & 51.21  & 41.24 & \textbf{61.39} & 57.7       & 53.05   \\ \hline
\end{tabular}
}
\quad 
\subfloat[MAE]{%
\begin{tabular}{|l|l|l|l|l|l|l|}
\hline
\multicolumn{1}{|l|}{}                            & SIS    & OHS    & OKS   & TUG   & Chair Stand & Average \\ \hline
\multicolumn{7}{|c|}{ML Regressors}                  \\ \hline
LR        				 & 0      & 0      & 0     & 0.01  & 0          & 0       \\ \hline
ET     					 & 3.3    & 3.71   & -1.18 & 5.38  & 5.86       & 3.41    \\ \hline
RF   					 & 0.06   & 13.79  & 1.82  & 4.41  & 10.56      & 6.13    \\ \hline
KNN     				 & 0      & 0      & 0     & 0     & 0          & 0       \\ \hline
SVR 					 & 0      & -0.23  & 0.42  & -4.39 & -0.68      & -0.98   \\ \hline
\multicolumn{7}{|c|}{DL Regressors}                     \\ \hline
NODE                     & 28.11  & 66.62  & 23.4  & 38.65 & 32.39      & 37.83   \\ \hline
FT-Transformer           & \textbf{41.37}  & \textbf{83.13}  & \textbf{46.78} & \textbf{45.6}  & \textbf{61.83}      & \textbf{55.74}   \\ \hline
TabPFN                   & -4.57  & -68.29 & 46.42 & -8.52 & 22.5       & -2.49   \\ \hline
TabNet                   & 35.53  & 33.65  & 26.86 & 42.98 & 36.61      & 35.13  \\ \hline
\end{tabular}
}
\end{table}

\begin{table}[h]
\centering
\caption{Percentage improvement of multi-output regressors with respect to LR baseline for leave-one-week-out cross-validation. The best values for each outcome across all the regressors are highlighted in bold.}
\label{tab_percent_overLR_LOWO}
\setlength{\tabcolsep}{4pt}
\subfloat[MSE]{%
\begin{tabular}{|l|l|l|l|l|l|l|}
\hline
\multicolumn{1}{|l|}{}                            & SIS    & OHS   & OKS   & TUG   & Chair Stand & Average \\ \hline
\multicolumn{7}{|c|}{ML Regressors}              \\ \hline
LR        				 & 0      & 0     & 0     & 0     & 0          & 0       \\ \hline
ET     					 & 45.48  & 59.89 & 63.02 & 54.49 & 58.80      & 56.34   \\ \hline
RF  					 & 39.53  & 61.77 & 60.78 & 50.80 & 54.58      & 53.49   \\ \hline
KNN     				 & 35.76  & 39.38 & 46.02 & 34.71 & 23.85      & 35.94   \\ \hline
SVR 					 & 26.55  & 34.63 & 34.72 & 27.63 & 29.86      & 30.68   \\ \hline
\multicolumn{7}{|c|}{DL Regressors}                 \\ \hline
NODE                     & \textbf{95.73}  & \textbf{96.83} & \textbf{96.85} & \textbf{96.27} & \textbf{96.10}      & \textbf{96.36}   \\ \hline
FT-Transformer           & 87.53  & 88.43 & 93.23 & 90.98 & 91.05      & 90.24   \\ \hline
TabPFN                   & 78.17  & 86.69 & 90.12 & 79.63 & 89.13      & 84.75   \\ \hline
TabNet                   & -15.34 & 6.48  & 2.51  & 2.33  & 5.56       & 0.31    \\ \hline
\end{tabular}
}
\quad 
\subfloat[MAE]{%
\begin{tabular}{|l|l|l|l|l|l|l|}
\hline
\multicolumn{1}{|l|}{}                            & SIS    & OHS   & OKS   & TUG   & Chair Stand & Average \\ \hline
\multicolumn{7}{|c|}{ML Regressors}              \\ \hline
LR        				 & 0      & 0     & 0     & 0     & 0          & 0       \\ \hline
ET     					 & 31.40  & 43.10 & 43.75 & 47.85 & 41.98      & 41.62   \\ \hline
RF  					 & 27.83  & 43.48 & 44.37 & 47.39 & 40.67      & 40.75   \\ \hline
KNN     				 & 32.64  & 34.08 & 38.22 & 42.87 & 27.39      & 35.04   \\ \hline
SVR 					 & 15.49  & 22.02 & 22.40 & 14.40 & 19.71      & 18.80   \\ \hline
\multicolumn{7}{|c|}{DL Regressors}                 \\ \hline
NODE                     & \textbf{89.66}  & \textbf{91.34} & \textbf{91.57} & \textbf{90.10} & \textbf{90.63}      & \textbf{90.66}   \\ \hline
FT-Transformer           & 77.74  & 80.90 & 84.02 & 79.75 & 81.50      & 80.78   \\ \hline
TabPFN                   & 59.06  & 68.41 & 72.23 & 59.93 & 70.85      & 66.10   \\ \hline
TabNet                   & -8.72  & 6.12  & 2.38  & 0.95  & 3.64       & 0.87    \\ \hline
\end{tabular}
}
\end{table}

\subsection{Leave-one-week-out cross-validation}
To validate the models' temporal robustness against week-to-week behavioural shifts, we extended our analysis to the leave-one-week-out cross-validation approach. By predicting variables across a held-out and unseen week, we can ascertain whether the multi-output regression successfully captures persistent clinical correlations over time, rather than merely memorizing time-specific routines. 
Consistent with the findings from the leave-one-person-out evaluation, the leave-one-week-out results establish the advantage of multi-output DL models for temporal prediction. 
The performance improvement of multi-output over single-output was found to be statistically significant (p-value = $0.003$) under the Wilcoxon signed-rank test.

The MSE and MAE for this validation are presented in Table \ref{tab_LOWO}. The error analysis demonstrates that traditional ML baselines largely failed to capitalize on the multi-output structure across temporal shifts. 
The multi-output LR model yielded an average MSE of 66.52 (Table \ref{tab_LOWO}), which was identical to its single-output counterpart. Conversely, tabular DL models were able to leverage the correlation between functional mobility and social isolation. The multi-output FT-Transformer showed the most improvement compared to the single-output. In a single-output setup, the FT-Transformer struggled with an average MSE of 26.01, while simultaneously predicting all five clinical variables drastically reduced the multi-output MSE to just 6.05. 
The NODE architecture remained the best-performing model in a leave-one-week-out cross-validation scenario as well, maintaining the lowest errors compared to other models. As can be seen in Table \ref{tab_LOWO}, compared to single-output, the multi-output regression allowed NODE to improve its performance from an MSE of 4.8 to 2.38. The improved performance of DL-based regressors is further supported by the MAE metrics. The multi-output NODE regressor achieved a low average MAE of 0.50 (down from 0.88 in single-output). Similarly, the FT-Transformer's average MAE dropped from 2.81 to 1.01.

To explicitly quantify the relative performance gains of the multi-output regression, Table \ref{tab_percent_LOWO} details the percentage improvements of multi-output over single-output regressors.
Tree-based models exhibited only marginal gains. The ET and RF regressors achieved modest average MSE improvements of 6.82\% and 11.37\%, respectively.
Conversely, the advanced tabular DL models successfully exploited interrelated temporal variations, resulting in notable percentage gains.
The FT-Transformer showed the most pronounced relative improvement, registering a 63.51\% overall MSE improvement and a 55.74\% MAE improvement. Further, FT-Transformer's ability to map week-to-week correlations resulted in a remarkable 91.76\% MSE improvement for the OHS and a 63.53\% improvement for the Chair Stand test relative to its single-output baseline.
NODE also demonstrated notable relative improvements, yielding a 44.73\% overall average MSE improvement and an average MAE improvement of 37.83\%. Specifically, NODE’s multi-output framework yielded a 74.17\% improvement in predicting OHS and a 45.83\% improvement for the TUG score over its single-output counterpart.
Other DL architectures like TabNet also saw immense benefits from the multi-target context, registering a 53.05\% average MSE improvement. TabPFN was an outlier that struggled with the temporal distribution shifts present in the leave-one-week-out scheme compared to the leave-one-person-out scheme; it averaged only a 2.81\% MSE improvement overall and even showed significantly degraded performance (-80.43\%) on the OHS variable. This isolated percentage analysis confirms that advanced multi-output DL models are uniquely equipped to leverage correlated temporal variations across unseen weeks, drastically outperforming isolated predictive approaches.

To quantify the performance of the advanced models under temporal distribution shifts, we analyzed the percentage improvement of all multi-output regressors against the multi-output LR baseline in Table \ref{tab_percent_overLR_LOWO}. By establishing the LR model as the baseline, we can analyze how well complex models decode the non-linear, temporal variations linking functional decline and social isolation across unseen weeks.
Traditional ML models demonstrated a much stronger ability to outperform the linear baseline compared to their performance in the leave-one-week-out cross-validation scenario. Tree-based ensemble regressors, such as ET and RF, achieved notable average MSE improvements of 56.34\% and 53.49\%, respectively. The KNN and SVR models also demonstrated moderate success, yielding average MSE improvements of 35.94\% and 30.68\% over the LR baseline.
The tabular DL models once again performed better, with the NODE model achieving a notable 96.36\% average MSE improvement across all five clinical variables. The performance of NODE  was stable across individual outcomes, yielding a 96.85\% MSE improvement for the OKS and a 96.83\% improvement for the OHS.
The FT-Transformer showed an average MSE improvement of 90.24\%, which included a 93.23\% improvement on the OKS and a 91.05\% improvement on the Chair Stand test. TabPFN proved promising as well, showing a 84.75\% average MSE improvement. However, TabNet was an outlier among the DL models; it struggled with temporal generalization against the linear baseline, resulting in a negligible average MSE improvement of 0.31\% and actually degrading by -15.34\% when predicting the SIS.
Further, the performance gain of DL models is corroborated by the MAE improvements. NODE maintained its best performance with an average MAE improvement of 90.66\%, followed closely by the FT-Transformer at 80.78\% and TabPFN at 66.10\%.

\begin{table*}
\centering
\caption{SHAP-based ranking of features for NODE multi-output regressor for leave-one-person-out cross-validation.}
\label{tab_SHAPrank}
\setlength{\tabcolsep}{4.5pt}
\begin{tabular}{|l|l|l|l|l|l|l|l|l|l|l|l|l|l|l|l|l|l|l|l|}
\hline
\diagbox[width=4.8cm, height=0.5cm]{Feature}{\#Fold}                               & 1  & 2  & 3  & 4  & 5  & 6  & 7  & 8  & 9  & 10 & 11 & 12 & 13 & 14 & 15 & 16 & 17 & 18 & Average \\ \hline
motion-max-timestamp                  & 12 & 17 & 6  & 1  & 2  & 6  & 1  & 1  & 3  & 3  & 2  & 3  & 4  & 1  & 3  & 4  & 2  & 3  & 4.11    \\ \hline
acceleration-minutes-with-data        & 3  & 13 & 4  & 6  & 11 & 3  & 7  & 3  & 7  & 2  & 6  & 8  & 9  & 5  & 6  & 11 & 11 & 5  & 6.67    \\ \hline
sleep-deep                            & 30 & 25 & 11 & 12 & 9  & 2  & 6  & 10 & 4  & 4  & 5  & 1  & 1  & 2  & 1  & 2  & 5  & 7  & 7.61    \\ \hline
acceleration-hours-with-data          & 2  & 12 & 7  & 10 & 12 & 5  & 14 & 6  & 8  & 5  & 8  & 10 & 10 & 9  & 8  & 17 & 16 & 6  & 9.17    \\ \hline
heartrate-hours-with-data             & 14 & 8  & 9  & 3  & 21 & 11 & 9  & 4  & 17 & 14 & 32 & 9  & 6  & 4  & 13 & 6  & 3  & 10 & 10.72   \\ \hline
motion-max                            & 20 & 24 & 21 & 15 & 3  & 13 & 5  & 15 & 21 & 23 & 13 & 4  & 7  & 3  & 4  & 1  & 1  & 1  & 10.78   \\ \hline
sleep-rem                             & 24 & 34 & 32 & 19 & 16 & 7  & 11 & 7  & 1  & 8  & 7  & 12 & 3  & 7  & 2  & 3  & 8  & 8  & 11.61   \\ \hline
acceleration-coefficient-of-variation & 25 & 6  & 3  & 4  & 14 & 9  & 2  & 14 & 9  & 11 & 1  & 18 & 11 & 15 & 18 & 8  & 17 & 25 & 11.67   \\ \hline
sleep-duration-to-wakeup              & 44 & 41 & 24 & 16 & 22 & 4  & 8  & 5  & 5  & 1  & 9  & 2  & 5  & 8  & 5  & 5  & 7  & 4  & 11.94   \\ \hline
acceleration-std                      & 28 & 7  & 2  & 7  & 15 & 10 & 3  & 16 & 12 & 12 & 3  & 19 & 13 & 16 & 20 & 9  & 18 & 27 & 13.17   \\ \hline
motion-ratio                          & 11 & 5  & 15 & 2  & 5  & 22 & 12 & 18 & 10 & 17 & 11 & 6  & 20 & 19 & 7  & 28 & 32 & 11 & 13.94   \\ \hline
sleep-heartrate-min                   & 6  & 14 & 16 & 9  & 7  & 16 & 4  & 12 & 14 & 6  & 12 & 13 & 14 & 6  & 29 & 26 & 23 & 29 & 14.22   \\ \hline
sleep-light                           & 8  & 33 & 17 & 13 & 13 & 15 & 10 & 8  & 15 & 18 & 4  & 5  & 16 & 23 & 14 & 16 & 22 & 9  & 14.39   \\ \hline
step-ratio                            & 13 & 9  & 14 & 5  & 25 & 12 & 17 & 9  & 18 & 19 & 40 & 11 & 8  & 13 & 17 & 7  & 9  & 15 & 14.50   \\ \hline
sleep-snoring                         & 37 & 39 & 28 & 25 & 28 & 26 & 20 & 23 & 2  & 9  & 14 & 7  & 2  & 12 & 9  & 10 & 13 & 2  & 17.00   \\ \hline
sleep-heartrate-mean                  & 4  & 16 & 20 & 8  & 8  & 8  & 24 & 20 & 30 & 7  & 16 & 31 & 33 & 14 & 16 & 18 & 24 & 21 & 17.67   \\ \hline
acceleration-movement-events-00to06   & 23 & 29 & 18 & 23 & 19 & 1  & 16 & 11 & 19 & 26 & 18 & 28 & 12 & 10 & 22 & 13 & 19 & 14 & 17.83   \\ \hline
sleep-duration-to-sleep               & 27 & 30 & 37 & 28 & 23 & 19 & 23 & 21 & 11 & 10 & 15 & 14 & 22 & 17 & 11 & 12 & 14 & 16 & 19.44   \\ \hline
sleep-wakeup-count                    & 17 & 31 & 22 & 17 & 17 & 23 & 19 & 2  & 31 & 16 & 22 & 17 & 27 & 25 & 15 & 15 & 25 & 13 & 19.67   \\ \hline
acceleration-mean                     & 42 & 35 & 1  & 32 & 39 & 20 & 26 & 31 & 6  & 13 & 10 & 16 & 28 & 20 & 10 & 14 & 10 & 22 & 20.83   \\ \hline
sleep-total                           & 26 & 27 & 13 & 11 & 24 & 35 & 13 & 13 & 13 & 25 & 17 & 20 & 35 & 29 & 12 & 21 & 33 & 12 & 21.06   \\ \hline
acceleration-entropy                  & 1  & 11 & 25 & 27 & 20 & 21 & 15 & 29 & 25 & 22 & 20 & 25 & 32 & 21 & 25 & 19 & 29 & 35 & 22.33   \\ \hline
acceleration-intradaily-variability   & 18 & 18 & 19 & 21 & 18 & 17 & 18 & 22 & 24 & 24 & 19 & 24 & 17 & 28 & 28 & 31 & 31 & 28 & 22.50   \\ \hline
step-max-timestamp                    & 10 & 10 & 26 & 20 & 26 & 24 & 21 & 17 & 26 & 30 & 35 & 26 & 29 & 24 & 19 & 24 & 21 & 20 & 22.67   \\ \hline
position-duration                     & 16 & 26 & 30 & 31 & 10 & 30 & 29 & 19 & 22 & 28 & 21 & 23 & 23 & 11 & 24 & 29 & 15 & 33 & 23.33   \\ \hline
heartrate-mean                        & 9  & 20 & 29 & 14 & 29 & 14 & 34 & 27 & 27 & 15 & 26 & 34 & 25 & 42 & 31 & 32 & 6  & 30 & 24.67   \\ \hline
sleep-heartrate-max                   & 7  & 21 & 27 & 24 & 30 & 18 & 37 & 28 & 33 & 20 & 24 & 32 & 21 & 34 & 27 & 20 & 28 & 31 & 25.67   \\ \hline
acceleration-movement-events-06to12   & 19 & 42 & 31 & 29 & 34 & 25 & 22 & 25 & 20 & 35 & 27 & 29 & 19 & 22 & 23 & 25 & 20 & 24 & 26.17   \\ \hline
acceleration-movement-events-18to24   & 41 & 23 & 35 & 34 & 33 & 28 & 30 & 24 & 23 & 29 & 44 & 15 & 15 & 27 & 21 & 22 & 12 & 17 & 26.28   \\ \hline
heartrate-std                         & 34 & 32 & 33 & 33 & 27 & 31 & 36 & 30 & 29 & 21 & 43 & 22 & 18 & 32 & 26 & 33 & 4  & 18 & 27.89   \\ \hline
position-count                        & 21 & 36 & 36 & 38 & 6  & 33 & 27 & 26 & 28 & 31 & 25 & 30 & 26 & 18 & 32 & 27 & 27 & 36 & 27.94   \\ \hline
heartrate-count                       & 22 & 3  & 34 & 39 & 35 & 27 & 38 & 34 & 32 & 27 & 28 & 27 & 24 & 33 & 33 & 37 & 30 & 32 & 29.72   \\ \hline
heartrate-max                         & 15 & 15 & 5  & 30 & 32 & 29 & 35 & 42 & 41 & 40 & 41 & 35 & 31 & 40 & 30 & 35 & 34 & 19 & 30.50   \\ \hline
acceleration-kurtosis                 & 45 & 45 & 43 & 22 & 37 & 40 & 25 & 38 & 16 & 33 & 23 & 21 & 30 & 30 & 35 & 30 & 35 & 26 & 31.89   \\ \hline
motion-count                          & 43 & 43 & 38 & 40 & 1  & 42 & 32 & 35 & 40 & 36 & 29 & 33 & 34 & 35 & 34 & 34 & 38 & 34 & 34.50   \\ \hline
step-mean                             & 36 & 4  & 10 & 35 & 38 & 38 & 43 & 33 & 35 & 38 & 36 & 41 & 43 & 39 & 38 & 36 & 41 & 37 & 34.50   \\ \hline
heartrate-min                         & 5  & 22 & 8  & 18 & 31 & 32 & 45 & 41 & 45 & 32 & 46 & 45 & 45 & 46 & 45 & 45 & 45 & 45 & 35.61   \\ \hline
motion-mean                           & 39 & 44 & 41 & 41 & 4  & 41 & 42 & 45 & 44 & 45 & 34 & 38 & 44 & 31 & 40 & 23 & 26 & 23 & 35.83   \\ \hline
step-max                              & 38 & 2  & 12 & 37 & 36 & 39 & 44 & 36 & 37 & 39 & 38 & 40 & 42 & 44 & 41 & 38 & 42 & 39 & 35.78   \\ \hline
step-count                            & 40 & 1  & 23 & 36 & 40 & 37 & 41 & 39 & 39 & 41 & 39 & 42 & 41 & 45 & 43 & 41 & 44 & 40 & 37.33   \\ \hline
acceleration-count                    & 35 & 40 & 45 & 44 & 41 & 34 & 28 & 37 & 36 & 34 & 31 & 39 & 37 & 36 & 37 & 40 & 39 & 42 & 37.50   \\ \hline
acceleration-skew                     & 29 & 38 & 40 & 26 & 42 & 43 & 33 & 44 & 34 & 43 & 30 & 36 & 36 & 38 & 44 & 39 & 43 & 38 & 37.56   \\ \hline
acceleration-sum                      & 31 & 37 & 44 & 45 & 43 & 36 & 31 & 40 & 38 & 37 & 33 & 43 & 40 & 37 & 39 & 42 & 40 & 43 & 38.83   \\ \hline
acceleration-movement-events-12to18   & 32 & 19 & 42 & 42 & 45 & 45 & 39 & 46 & 43 & 44 & 37 & 37 & 39 & 41 & 36 & 44 & 37 & 41 & 39.39   \\ \hline
acceleration-movement-events-24h      & 33 & 28 & 39 & 43 & 44 & 44 & 40 & 43 & 42 & 42 & 45 & 44 & 38 & 43 & 42 & 43 & 36 & 44 & 40.72   \\ \hline
position-distance-travelled           & 46 & 46 & 46 & 46 & 46 & 46 & 46 & 32 & 46 & 46 & 42 & 46 & 46 & 26 & 46 & 46 & 46 & 46 & 43.89   \\ \hline
\end{tabular}
\end{table*}

\begin{figure*}[!ht]
\centering
\stackunder[2pt]{\includegraphics[width=\textwidth]{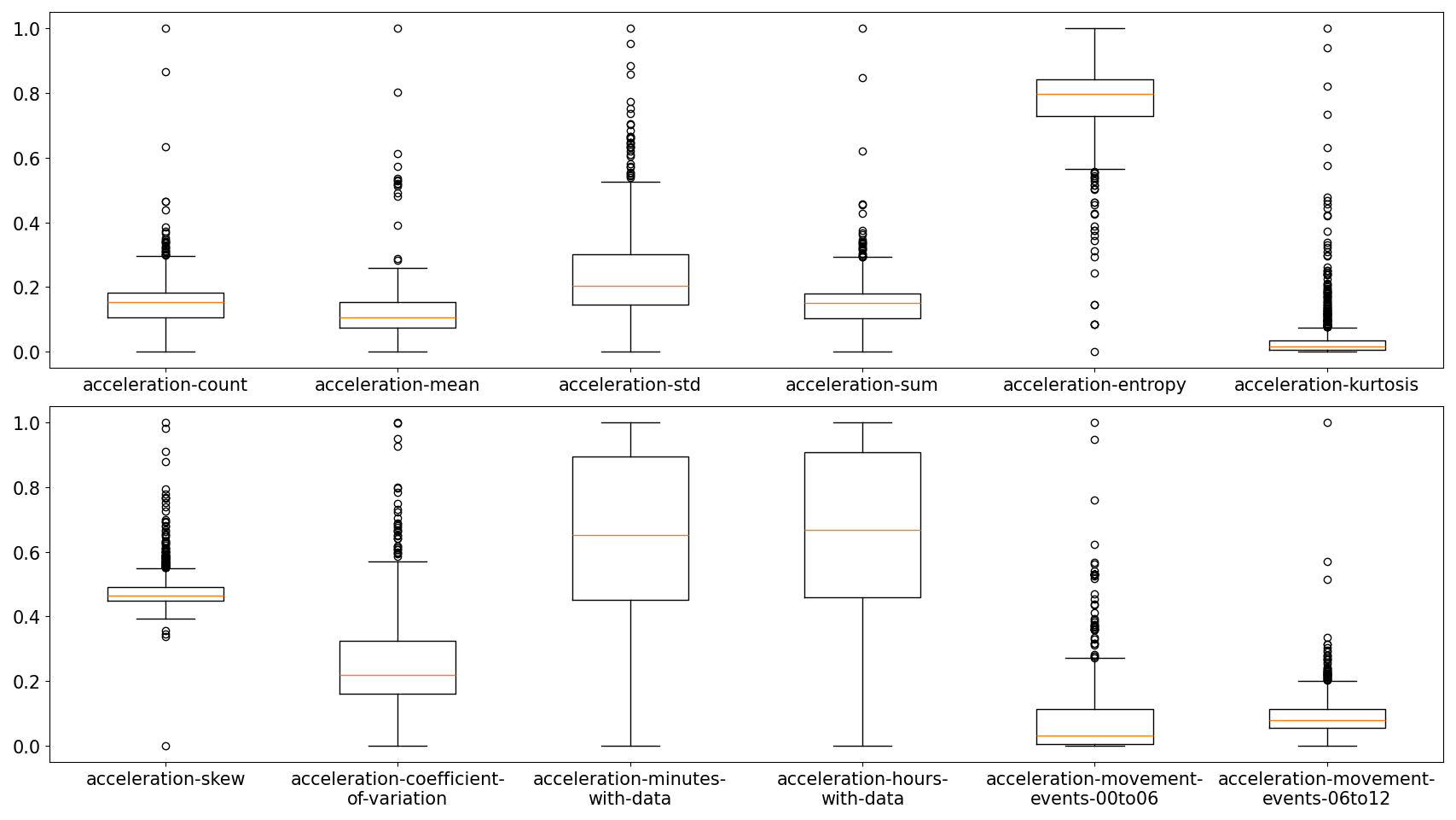}}{}
\stackunder[2pt]{\includegraphics[width=\textwidth]{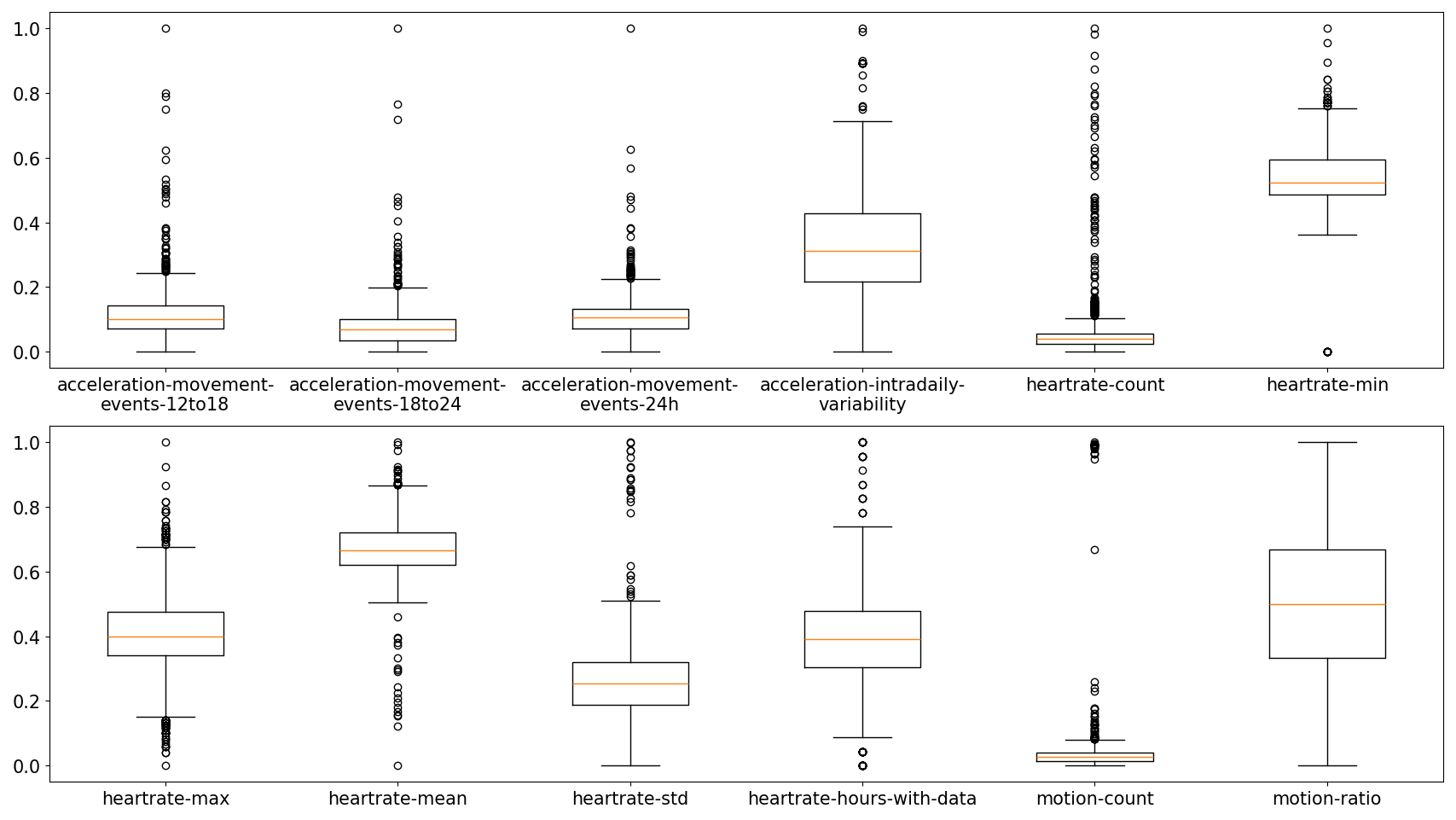}}{}
\caption{Outlier feature plots.}
\label{fig_outlier_1}
\end{figure*}

\begin{figure*}[!ht]
\centering
\stackunder[2pt]{\includegraphics[width=\textwidth]{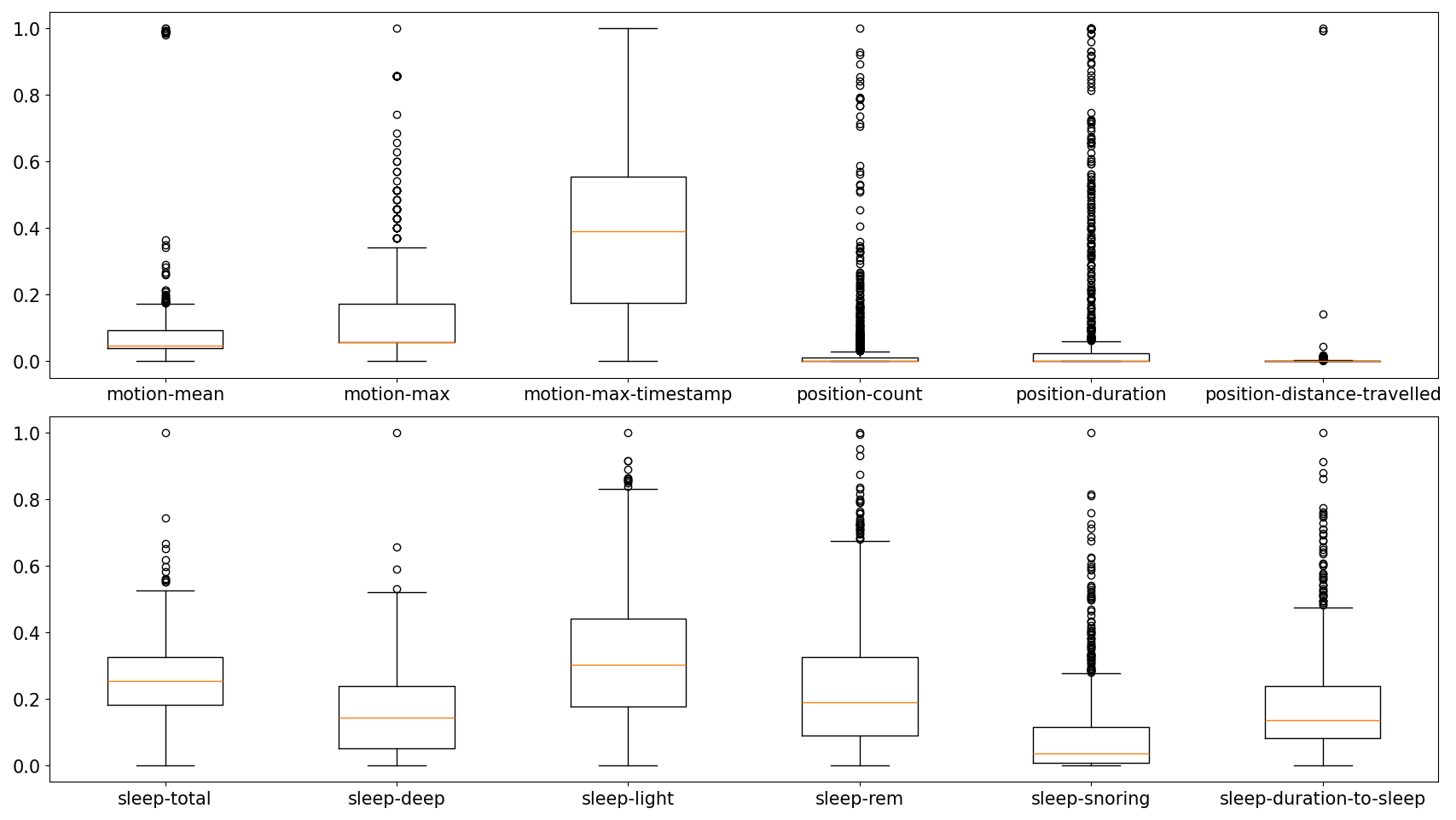}}{}
\stackunder[2pt]{\includegraphics[width=\textwidth]{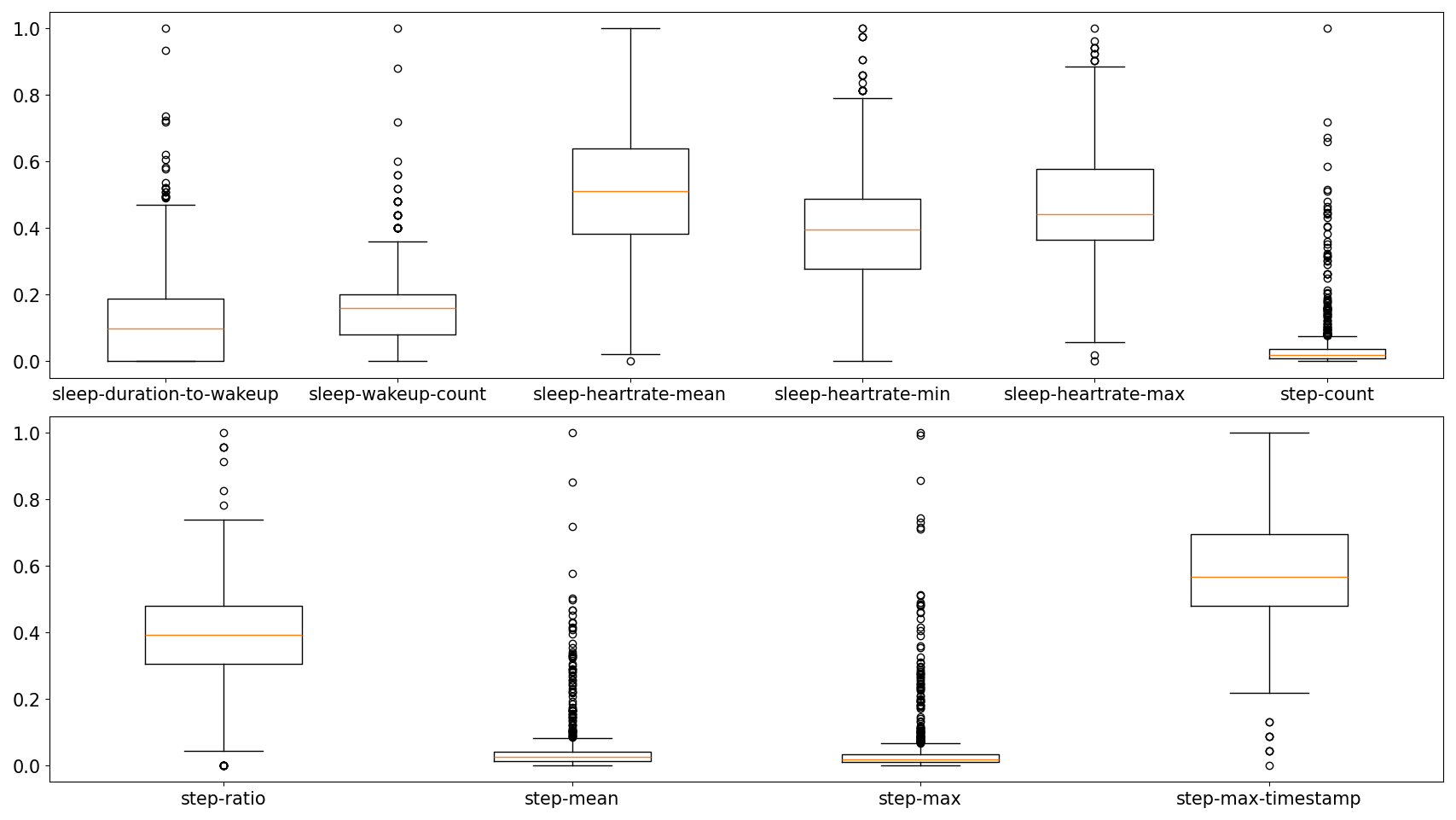}}{}
\caption{Outlier feature plots (continued).}
\label{fig_outlier_2}
\end{figure*}

\subsection{Features Distribution Analysis}
To understand the challenges associated with modelling continuous wearable sensor data, it is important to analyze the distribution and inherent outliers across the extracted feature set. Figures \ref{fig_outlier_1} and \ref{fig_outlier_2} illustrate the normalized box plots for the diverse physiological and behavioural features, encompassing acceleration, heart rate, motion, position, sleep, and step metrics. Statistical outliers may capture acute behavioural deviations, such as sudden periods of extreme restlessness, anomalous sleep fragmentation, or rare excursions outside the home.

\begin{itemize}
    \item The acceleration and motion modalities exhibit significant right-skewed distributions with dense clusters of upper outliers. Features such as acceleration-kurtosis, acceleration-movement-events-00to06, and motion-count are heavily compressed near zero but feature long tails extending towards 1.0. This distribution pattern reflects prolonged periods of sedentary behaviour with bursts of physical activity or nighttime restlessness. Similarly, the step-count, step-mean, and step-max features display massive upper outlier densities.
    \item The mobility of the participants, captured via GPS positioning, provides the evidence of social isolation risk. The position-duration and position-distance-travelled features, along with position-count, possess median values clustered at the absolute bottom of the distribution space, with almost all variations registering as extreme upper outliers. This indicates that the community-dwelling older adults in this study spent the vast majority of their time at home. As reduced mobility can be a primary driver of social isolation, these highly skewed positional outliers can be the vital features for mapping the SIS.
    \item Heart rate features such as heartrate-max contain pronounced upper outliers, indicating episodic cardiac exertion. While the features like sleep-total and sleep-light follow relatively normal, broad distributions, sleep-duration-to-sleep and sleep-wakeup-count features, along with sleep-snoring, have extreme upper outliers. These upper-bound data points can represent nights of disrupted rest.
\end{itemize}

The severe skew and heavy outlier presence in features can be used to explain the relatively lower performance of ML models compared to DL models. Standard LR model inherently fails to map long-tailed distributions, while DL transformer models utilize attention mechanisms. They can process non-normally distributed, tabular features, ensuring that such features can be appropriately leveraged to predict functional decline and social isolation.

\subsection{Features Importance and Interpretability}
To understand the decision-making process of the best-performing NODE regressor and extract actionable clinical insights, we utilized SHapley Additive exPlanations (SHAP) \cite{lundberg2017unified} to evaluate feature importance. Table \ref{tab_SHAPrank} details the SHAP-based ranking of $46$ distinct physiological and behavioural features for the multi-output NODE regressor across all 18 folds of the leave-one-person-out cross-validation. By analyzing the average rankings, we can identify which features are most critical for simultaneously predicting social isolation and functional decline in unseen older adults (across all folds).

The SHAP analysis reveals that the timing of behaviour is significantly more predictive than the raw volume of physical activity. The most important feature across the entire dataset is motion-max-timestamp, achieving the highest overall average rank of 4.11. This indicates that the specific hour of the day when an older adult achieves their maximum motion is a vital predictor of their physical and social health.
Similarly, sleep-based features heavily influence the model's predictive capabilities. The duration of deep sleep (sleep-deep) ranked as the third most important feature overall (average rank 7.61), while sleep-rem (average rank 11.61) and sleep-duration-to-wakeup (average rank 11.94) also ranked within the top ten most impactful predictors.

The SHAP rankings show the limitations of relying on traditional, raw volumetric metrics for clinical prediction. Activity counters such as step-count (average rank 37.33), acceleration-sum (average rank 38.83), and acceleration-movement-events-24h (average rank 40.72) ranked near the bottom of the feature set. This demonstrates that merely knowing how much an older adult moves within 24 hours provides very little predictive value for complex clinical outcomes compared to knowing when they move and the variance of that movement (e.g., acceleration-coefficient-of-variation, average rank 11.67). Further, position-distance-travelled emerged as the absolute least important feature for the multi-output NODE model, ranking last with an average placement of 43.89. As established in the data distribution analysis, the vast majority of the community-dwelling cohort spent their time at home, rendering outside positional variance too sparse to serve as a reliable, generalized predictor across individual folds.

\section{Conclusion and Future Work}
This paper presented a new multi-output regression formulation of predicting multiple clinical scores related to functional mobility and social isolation among older adults recovering in the community post lower-limb surgery. The multi-output regression models were trained on multimodal sensor data collected over 1008 days from 18 patients in a leave-one-participant-out and leave-one-week-out cross-validation evaluation framework. Several key findings were made in this analysis -- multi-output regressors performed better than single-output regressors for all the ML and DL approaches tested; DL approaches performed better than ML methods, with NODE method emerging as the best multi-output regressor; feature importance suggested the key role of mobility and physiological data in patients' functional recovery and social isolation. 

There are several limitations of this work, the first being a small sample size. Although 1008 days worth of multimodal sensor data was analyzed, it was only collected from 18 patients. Moreover, the female participants dominated the cohort; however, the sample size is quite small to perform any sex-based analysis and draw conclusions. A large-scale data collection using the MAISON system could reveal general population-wide, sex/gender, and other ethno-demographic intricacies in this population. The dataset was collected from one geographic region (Greater Toronto Area) and included both summer and winter seasons, which could be a confounding factor because the winter season can be very harsh and could restrict mobility in this population. Clinical scores were collected every two weeks but assigned to each of the 14 daily sensor records preceding the assessment. This makes the 1,008 participant-days appear like independent observations when the true independent sample is only 18 participants with a limited number of assessment occasions.

This analysis opens several new directions both in terms of clinical and technical perspectives. A longitudinal analysis for each participant can help build an individual recovery profile, which can be modelled through sequential DL models, such as transformers and recurrent neural networks. The current feature importance analysis provides semi-contextual information per feature; however, it lacks descriptive analysis. Advanced modelling techniques with Large language models \cite{khan2025explaining} can be used to describe the phenotypes, symptoms, and potential causes of patients high or low recovery and social engagements in the community. The next version of this dataset that contains GPS coordinates (GEOFRAIL dataset \cite{abedi2026longitudinal}) could be a relevant resource to include geographical information (e.g., public parks) and augment it with neighborhood demographics (crime rate, average income) to bring more contextual information to these models besides modeling only on the objective sensor data.

\section*{Acknowledgment}
This study is funded by Liwa University, United Arab Emirates (Grant: IRG-ENG-003-2025).

\bibliographystyle{IEEEtran}
\bibliography{ms}

\begin{thebibliography}{10}
\providecommand{\url}[1]{#1}
\csname url@samestyle\endcsname
\providecommand{\newblock}{\relax}
\providecommand{\bibinfo}[2]{#2}
\providecommand{\BIBentrySTDinterwordspacing}{\spaceskip=0pt\relax}
\providecommand{\BIBentryALTinterwordstretchfactor}{4}
\providecommand{\BIBentryALTinterwordspacing}{\spaceskip=\fontdimen2\font plus
\BIBentryALTinterwordstretchfactor\fontdimen3\font minus
  \fontdimen4\font\relax}
\providecommand{\BIBforeignlanguage}[2]{{%
\expandafter\ifx\csname l@#1\endcsname\relax
\typeout{** WARNING: IEEEtran.bst: No hyphenation pattern has been}%
\typeout{** loaded for the language `#1'. Using the pattern for}%
\typeout{** the default language instead.}%
\else
\language=\csname l@#1\endcsname
\fi
#2}}
\providecommand{\BIBdecl}{\relax}
\BIBdecl

\bibitem{ishaku2025enhanced}
Z.~Ishaku, D.~I. Koshy, and M.~A. Bala, ``Enhanced recovery after surgery
  (eras) pathways in elective total joint arthroplasty,'' \emph{Cureus},
  vol.~17, no.~9, pp. e91\,481--e91\,481, 2025.

\bibitem{singh2025rehabilitation}
A.~Singh, A.~Kumar, S.~Y. Kale, S.~Prakash, and V.~Kumar, ``Rehabilitation
  after lower limb fracture fixation in osteoporotic bone,'' \emph{Indian
  Journal of Orthopaedics}, vol.~59, no.~3, pp. 405--413, 2025.

\bibitem{mohammed2025psychosocial}
U.~F. Mohammed, E.~Tornu, and L.~Aziato, ``Psychosocial experiences of adults
  with lower limb fracture in ghana: A qualitative study,'' \emph{Journal of
  Patient Experience}, vol.~12, p. 23743735251400003, 2025.

\bibitem{shear2025predicting}
B.~M. Shear, D.~J. Brodke, G.~R. Hancock, P.~McGlone, H.~Demyanovich, V.~Li,
  A.~Bell, D.~Okhuereigbe, G.~P. Slobogean, R.~V. O’Toole \emph{et~al.},
  ``Predicting post-fracture recovery with smartphone mobility data: a
  proof-of-concept study,'' \emph{The Journal of bone and joint surgery.
  American volume}, vol. 107, no.~11, p. e57, 2025.

\bibitem{rao2026management}
H.~Rao, L.~Luo, J.~Cheng, G.~Wu, H.~Zhang, X.~Zhang, J.~Qin, and Z.~Wang,
  ``Management of hip fracture in older adults with cognitive impairment: a
  narrative review,'' \emph{Frontiers in Public Health}, vol.~14, p. 1816268,
  2026.

\bibitem{mangione2005can}
K.~K. Mangione, R.~L. Craik, S.~S. Tomlinson, and K.~M. Palombaro, ``Can
  elderly patients who have had a hip fracture perform moderate-to
  high-intensity exercise at home?'' \emph{Physical therapy}, vol.~85, no.~8,
  pp. 727--739, 2005.

\bibitem{bevilacqua2021association}
G.~Bevilacqua, K.~A. Jameson, J.~Zhang, I.~Bloom, K.~A. Ward, C.~Cooper, and
  E.~M. Dennison, ``The association between social isolation and
  musculoskeletal health in older community-dwelling adults: findings from the
  hertfordshire cohort study,'' \emph{Quality of life research}, vol.~30,
  no.~7, pp. 1913--1924, 2021.

\bibitem{mandl2024effect}
L.~A. Mandl, M.~Rajan, R.~A. Lipschultz, S.~Lian, D.~Sheira, M.~B. Frey, Y.~M.
  Shea, and J.~M. Lane, ``The effect of social isolation on 1-year outcomes
  after surgical repair of low-energy hip fracture,'' \emph{Journal of
  orthopaedic trauma}, vol.~38, no.~4, pp. e149--e156, 2024.

\bibitem{abedi2022maison}
A.~Abedi, F.~Dayyani, C.~Chu, and S.~S. Khan, ``Maison-multimodal ai-based
  sensor platform for older individuals,'' in \emph{2022 IEEE International
  Conference on Data Mining Workshops (ICDMW)}.\hskip 1em plus 0.5em minus
  0.4em\relax IEEE, 2022, pp. 238--242.

\bibitem{abedi2025multimodal}
A.~Abedi, C.~H. Chu, and S.~S. Khan, ``Multimodal sensor dataset for monitoring
  older adults post lower limb fractures in community settings,''
  \emph{Scientific Data}, vol.~12, no.~1, p. 733, 2025.

\bibitem{maison-llf}
\BIBentryALTinterwordspacing
A.~Abedi, C.~Chu, and S.~S. Khan, ``Maison-llf: Multimodal sensor dataset for
  monitoring older adults post lower-limb fractures in community settings,''
  2025, accessed on: June 22, 2026. [Online]. Available:
  \url{https://zenodo.org/records/17943110}
\BIBentrySTDinterwordspacing

\bibitem{dayyani2024correlations}
F.~Dayyani, C.~H. Chu, A.~Abedi, and S.~S. Khan, ``Correlations between social
  isolation and functional decline in older adults after lower limb fractures
  using multimodal sensors: A pilot study,'' \emph{Algorithms}, vol.~17, no.~9,
  p. 383, 2024.

\bibitem{austin2016smart}
J.~Austin, H.~H. Dodge, T.~Riley, P.~G. Jacobs, S.~Thielke, and J.~Kaye, ``A
  smart-home system to unobtrusively and continuously assess loneliness in
  older adults,'' \emph{IEEE journal of translational engineering in health and
  medicine}, vol.~4, pp. 1--11, 2016.

\bibitem{martinez2020automatic}
A.~Martinez~Rebollar, M.~Gonzalez~Mendoza, H.~Estrada~Esquivel,
  W.~Campos~Francisco, and V.~Campos~Ortiz, ``Automatic detection of social
  isolation based on human behavior analysis,'' \emph{Computaci{\'o}n y
  Sistemas}, vol.~24, no.~4, pp. 1527--1538, 2020.

\bibitem{goonawardene2017sensor}
N.~Goonawardene, X.~Toh, and H.-P. Tan, ``Sensor-driven detection of social
  isolation in community-dwelling elderly,'' in \emph{International Conference
  on Human Aspects of IT for the Aged Population}.\hskip 1em plus 0.5em minus
  0.4em\relax Springer, 2017, pp. 378--392.

\bibitem{khan2023sensor}
S.~S. Khan, T.~Gu, L.~Spinelli, and R.~H. Wang, ``Sensor-based assessment of
  social isolation in community-dwelling older adults: a scoping review,''
  \emph{Biomedical engineering online}, vol.~22, no.~1, p.~18, 2023.

\bibitem{hu2017elderly}
R.~Hu, H.~Pham, P.~Buluschek, and D.~Gatica-Perez, ``Elderly people living
  alone: Detecting home visits with ambient and wearable sensing,'' in
  \emph{Proceedings of the 2nd International Workshop on Multimedia for
  Personal Health and Health Care}, 2017, pp. 85--88.

\bibitem{schutz2021sensor}
N.~Sch{\"u}tz, A.~Botros, S.~B. Hassen, H.~Saner, P.~Buluschek, P.~Urwyler,
  B.~Pais, V.~Santschi, D.~Gatica-Perez, R.~M. M{\"u}ri \emph{et~al.}, ``A
  sensor-driven visit detection system in older adults’ homes: towards
  digital late-life depression marker extraction,'' \emph{IEEE Journal of
  Biomedical and Health Informatics}, vol.~26, no.~4, pp. 1560--1569, 2021.

\bibitem{walsh2014inferring}
L.~Walsh, A.~Kealy, J.~Loane, J.~Doyle, and R.~Bond, ``Inferring health metrics
  from ambient smart home data,'' in \emph{2014 IEEE International Conference
  on Bioinformatics and Biomedicine (BIBM)}.\hskip 1em plus 0.5em minus
  0.4em\relax IEEE, 2014, pp. 27--32.

\bibitem{fan2023digital}
S.~Fan, J.~Ye, Q.~Xu, R.~Peng, B.~Hu, Z.~Pei, Z.~Yang, and F.~Xu, ``Digital
  health technology combining wearable gait sensors and machine learning
  improve the accuracy in prediction of frailty,'' \emph{Frontiers in Public
  Health}, vol.~11, p. 1169083, 2023.

\bibitem{giggins2025unsupervised}
O.~M. Giggins, G.~Vavasour, and J.~Doyle, ``Unsupervised assessment of frailty
  status using wearable sensors: a feasibility study among community-dwelling
  older adults,'' \emph{Advances in Rehabilitation Science and Practice},
  vol.~14, p. 27536351241311845, 2025.

\bibitem{palermo2023tihm}
F.~Palermo, Y.~Chen, A.~Capstick, N.~Fletcher-Loyd, C.~Walsh, S.~Kouchaki,
  J.~True, O.~Balazikova, E.~Soreq, G.~Scott \emph{et~al.}, ``Tihm: An open
  dataset for remote healthcare monitoring in dementia,'' \emph{Scientific
  data}, vol.~10, no.~1, p. 606, 2023.

\bibitem{north2024predicting}
K.~North, G.~Simpson, W.~Geiger, A.~Cizik, D.~Rothberg, and R.~Hitchcock,
  ``Predicting the healing of lower extremity fractures using wearable ground
  reaction force sensors and machine learning,'' \emph{Sensors}, vol.~24,
  no.~16, p. 5321, 2024.

\bibitem{abedi2026longitudinal}
A.~Abedi, C.~H. Chu, and S.~S. Khan, ``A longitudinal geospatial multimodal
  dataset of post-discharge frailty, physiology, mobility, and neighborhoods,''
  \emph{arXiv preprint arXiv:2602.00060}, 2026.

\bibitem{khan2025explaining}
S.~S. Khan, A.~Abedi, and C.~H. Chu, ``Explaining recovery trajectories of
  older adults post lower-limb fracture using modality-wise multiview
  clustering and large language models,'' in \emph{International Conference on
  Big Data Analytics and Knowledge Discovery}.\hskip 1em plus 0.5em minus
  0.4em\relax Springer, 2025, pp. 271--285.

\bibitem{zhang2012multi}
D.~Zhang, D.~Shen, A.~D.~N. Initiative \emph{et~al.}, ``Multi-modal multi-task
  learning for joint prediction of multiple regression and classification
  variables in alzheimer's disease,'' \emph{NeuroImage}, vol.~59, no.~2, pp.
  895--907, 2012.

\bibitem{zhou2012modeling}
J.~Zhou, J.~Liu, V.~A. Narayan, and J.~Ye, ``Modeling disease progression via
  fused sparse group lasso,'' in \emph{Proceedings of the 18th ACM SIGKDD
  international conference on Knowledge discovery and data mining}, 2012, pp.
  1095--1103.

\bibitem{el2020multimodal}
S.~El-Sappagh, T.~Abuhmed, S.~R. Islam, and K.~S. Kwak, ``Multimodal multitask
  deep learning model for alzheimer’s disease progression detection based on
  time series data,'' \emph{Neurocomputing}, vol. 412, pp. 197--215, 2020.

\bibitem{harutyunyan2019multitask}
H.~Harutyunyan, H.~Khachatrian, D.~C. Kale, G.~Ver~Steeg, and A.~Galstyan,
  ``Multitask learning and benchmarking with clinical time series data,''
  \emph{Scientific data}, vol.~6, no.~1, p.~96, 2019.

\bibitem{shickel2021multi}
B.~Shickel, P.~J. Tighe, A.~Bihorac, and P.~Rashidi, ``Multi-task prediction of
  clinical outcomes in the intensive care unit using flexible multimodal
  transformers,'' \emph{arXiv preprint arXiv:2111.05431}, 2021.

\bibitem{chan2024multi}
T.~H. Chan, G.~Yin, K.~Bae, and L.~Yu, ``Multi-task heterogeneous graph
  learning on electronic health records,'' \emph{Neural Networks}, vol. 180, p.
  106644, 2024.

\bibitem{cui2018prediction}
L.~Cui, X.~Xie, Z.~Shen, R.~Lu, and H.~Wang, ``Prediction of the healthcare
  resource utilization using multi-output regression models,'' \emph{IISE
  Transactions on Healthcare Systems Engineering}, vol.~8, no.~4, pp. 291--302,
  2018.

\bibitem{ding2019effectiveness}
D.~Y. Ding, C.~Simpson, S.~Pfohl, D.~C. Kale, K.~Jung, and N.~H. Shah, ``The
  effectiveness of multitask learning for phenotyping with electronic health
  records data,'' in \emph{Pacific Symposium on Biocomputing. Pacific Symposium
  on Biocomputing}, vol.~24, 2019, p.~18.

\bibitem{cui2024automated}
S.~Cui and P.~Mitra, ``Automated multi-task learning for joint disease
  prediction on electronic health records,'' \emph{Advances in Neural
  Information Processing Systems}, vol.~37, pp. 129\,187--129\,208, 2024.

\bibitem{chen2020metier}
L.~Chen, Y.~Zhang, and L.~Peng, ``Metier: a deep multi-task learning based
  activity and user recognition model using wearable sensors,''
  \emph{Proceedings of the ACM on Interactive, Mobile, Wearable and Ubiquitous
  Technologies}, vol.~4, no.~1, pp. 1--18, 2020.

\bibitem{nisar2023hierarchical}
M.~A. Nisar, K.~Shirahama, M.~T. Irshad, X.~Huang, and M.~Grzegorzek, ``A
  hierarchical multitask learning approach for the recognition of activities of
  daily living using data from wearable sensors,'' \emph{Sensors}, vol.~23,
  no.~19, p. 8234, 2023.

\bibitem{duan2023multitask}
F.~Duan, T.~Zhu, J.~Wang, L.~Chen, H.~Ning, and Y.~Wan, ``A multitask deep
  learning approach for sensor-based human activity recognition and
  segmentation,'' \emph{IEEE Transactions on Instrumentation and Measurement},
  vol.~72, pp. 1--12, 2023.

\bibitem{khan2022treatment}
A.~Khan, A.~Hazart, O.~Galarraga, S.~Garcia-Salicetti, and V.~Vigneron,
  ``Treatment outcome prediction using multi-task learning: application to
  botulinum toxin in gait rehabilitation,'' \emph{Sensors}, vol.~22, no.~21, p.
  8452, 2022.

\bibitem{saylam2024multitask}
B.~Saylam and {\"O}.~D. {\.I}ncel, ``Multitask learning for mental health:
  depression, anxiety, stress (das) using wearables,'' \emph{Diagnostics},
  vol.~14, no.~5, p. 501, 2024.

\bibitem{kang2025exploring}
B.~Kang, M.~K. Park, J.~I. Kim, S.~Yoon, S.-J. Heo, C.~Kang, S.~Lee, Y.~Choi,
  and D.~Hong, ``Exploring factors related to social isolation among older
  adults in the predementia stage using ecological momentary assessments and
  actigraphy: Machine learning approach,'' \emph{Journal of Medical Internet
  Research}, vol.~27, p. e69379, 2025.

\bibitem{ji2026smile}
X.~Ji, A.~Yuh, V.~Erd{\'e}lyi, T.~Mizumoto, H.~Choi, S.~Harrison, E.~Cho,
  T.~Suehiro, T.~Nakagawa, Y.~Cheng \emph{et~al.}, ``Smile: Sensor data-driven
  detection and counterfactual pattern analysis for loneliness in older
  adults,'' \emph{ACM Transactions on Computing for Healthcare}, vol.~7, no.~2,
  pp. 1--30, 2026.

\bibitem{prabhu2022sensor}
D.~Prabhu, M.~Kholghi, M.~Sandhu, W.~Lu, K.~Packer, L.~Higgins, and
  D.~Silvera-Tawil, ``Sensor-based assessment of social isolation and
  loneliness in older adults: a survey,'' \emph{Sensors}, vol.~22, no.~24, p.
  9944, 2022.

\bibitem{qirtas2022loneliness}
M.~M. Qirtas, E.~Zafeiridi, D.~Pesch, and E.~B. White, ``Loneliness and social
  isolation detection using passive sensing techniques: scoping review,''
  \emph{JMIR mHealth and uHealth}, vol.~10, no.~4, p. e34638, 2022.

\bibitem{parraga2026sensor}
M.~M. P{\'a}rraga~Vico, J.~M. Morcillo~Mart{\'\i}nez, J.~F.
  Gait{\'a}n-Guerrero, J.~L. Herreros~B{\'o}dalo, M.~Espinilla~Est{\'e}vez, and
  J.~C. Cuevas~Mart{\'\i}nez, ``Sensor-based technologies for the detection of
  unwanted loneliness in older adults: A systematic review,'' \emph{Sensors},
  vol.~26, no.~7, p. 2028, 2026.

\bibitem{cihi2024proms}
{Canadian Institute for Health Information}, ``Hip and knee arthroplasty
  proms,'' 2024, available:
  https://www.cihi.ca/en/patient-reported-outcome-measures-proms/hip-and-knee-arthroplasty-proms.

\bibitem{beauchet2011timed}
O.~Beauchet, B.~Fantino, G.~Allali, S.~Muir, M.~Montero-Odasso, and
  C.~Annweiler, ``Timed up and go test and risk of falls in older adults: a
  systematic review,'' \emph{The journal of nutrition, health \& aging},
  vol.~15, no.~10, pp. 933--938, 2011.

\bibitem{jones199930}
C.~J. Jones, R.~E. Rikli, and W.~C. Beam, ``A 30-s chair-stand test as a
  measure of lower body strength in community-residing older adults,''
  \emph{Research quarterly for exercise and sport}, vol.~70, no.~2, pp.
  113--119, 1999.

\bibitem{philip2020social}
K.~E. Philip, M.~I. Polkey, N.~S. Hopkinson, A.~Steptoe, and D.~Fancourt,
  ``Social isolation, loneliness and physical performance in older-adults:
  fixed effects analyses of a cohort study,'' \emph{Scientific reports},
  vol.~10, no.~1, p. 13908, 2020.

\bibitem{zare2024social}
Z.~Zare, G.~Ghane, H.~Shahsavari, S.~Ahmadnia, and S.~Ghiyasvandian, ``Social
  life after hip fracture: a qualitative study,'' \emph{Journal of Patient
  Experience}, vol.~11, p. 23743735241241174, 2024.

\bibitem{nicholson2020psychometric}
N.~R. Nicholson~Jr, R.~Feinn, E.~Casey, and J.~Dixon, ``Psychometric evaluation
  of the social isolation scale in older adults,'' \emph{The Gerontologist},
  vol.~60, no.~7, pp. e491--e501, 2020.

\bibitem{wylde2005oxford}
V.~Wylde, I.~D. Learmonth, and V.~J. Cavendish, ``The oxford hip score: the
  patient's perspective,'' \emph{Health and quality of life outcomes}, vol.~3,
  no.~1, p.~66, 2005.

\bibitem{whitehouse2005oxford}
S.~L. Whitehouse, A.~W. Blom, A.~H. Taylor, G.~T. Pattison, and G.~C.
  Bannister, ``The oxford knee score; problems and pitfalls,'' \emph{The Knee},
  vol.~12, no.~4, pp. 287--291, 2005.

\bibitem{podsiadlo1991timed}
D.~Podsiadlo and S.~Richardson, ``The timed “up \& go”: a test of basic
  functional mobility for frail elderly persons,'' \emph{Journal of the
  American geriatrics Society}, vol.~39, no.~2, pp. 142--148, 1991.

\bibitem{chen2009normative}
H.-T. Chen, C.-H. Lin, and L.-H. Yu, ``Normative physical fitness scores for
  community-dwelling older adults,'' \emph{Journal of Nursing Research},
  vol.~17, no.~1, pp. 30--41, 2009.

\bibitem{popovneural}
S.~Popov, S.~Morozov, and A.~Babenko, ``Neural oblivious decision ensembles for
  deep learning on tabular data,'' in \emph{8th International Conference on
  Learning Representations, {ICLR} 2020, Addis Ababa, Ethiopia, April 26-30,
  2020}.\hskip 1em plus 0.5em minus 0.4em\relax OpenReview.net, 2020.

\bibitem{gorishniy2021revisiting}
Y.~Gorishniy, I.~Rubachev, V.~Khrulkov, and A.~Babenko, ``Revisiting deep
  learning models for tabular data,'' \emph{Advances in neural information
  processing systems}, vol.~34, pp. 18\,932--18\,943, 2021.

\bibitem{hollmanntabpfn}
N.~Hollmann, S.~M{\"{u}}ller, K.~Eggensperger, and F.~Hutter, ``Tabpfn: {A}
  transformer that solves small tabular classification problems in a second,''
  in \emph{The Eleventh International Conference on Learning Representations,
  {ICLR} 2023, Kigali, Rwanda, May 1-5, 2023}.\hskip 1em plus 0.5em minus
  0.4em\relax OpenReview.net, 2023.

\bibitem{arik2021tabnet}
S.~{\"O}. Arik and T.~Pfister, ``Tabnet: Attentive interpretable tabular
  learning,'' in \emph{Proceedings of the AAAI conference on artificial
  intelligence}, vol.~35, no.~8, 2021, pp. 6679--6687.

\bibitem{lundberg2017unified}
S.~M. Lundberg and S.-I. Lee, ``A unified approach to interpreting model
  predictions,'' \emph{Advances in neural information processing systems},
  vol.~30, 2017.

\end{thebibliography}

\begin{IEEEbiography}[{\includegraphics[width=1.1in,height=1.25in,clip]{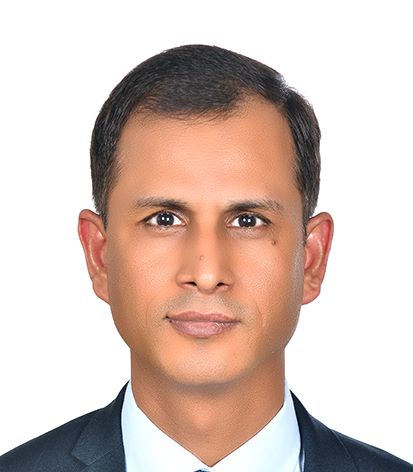}}]{Santosh Ray}, Ph.D., is Professor and Director of Institutional Research and Quality Assurance at Liwa University, UAE.  He has held senior academic leadership positions, including Acting President, Acting Vice President for Academic Affairs, Dean, and Head of Research, and has published extensively in his areas of expertise. 

His research interests encompass artificial intelligence, machine learning, data analytics, educational technology, big data applications, information retrieval, quality assurance in higher education, and institutional effectiveness. He has authored numerous scholarly publications and actively contributes to research, accreditation, and academic quality enhancement initiatives in higher education.
\end{IEEEbiography}

\begin{IEEEbiography}[{\includegraphics[width=1.1in,height=1.25in,clip]{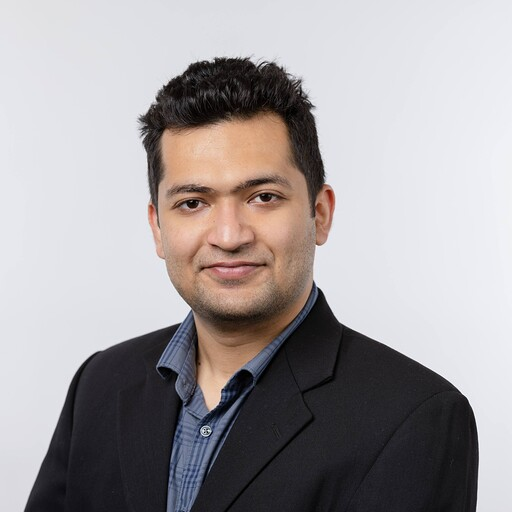}}]{Pratik K. Mishra} received the M.S. degree in computer science and engineering from the Indian Institute of Technology Indore, India, in 2020. He is currently working toward the Ph.D. degree with the Institute of Biomedical Engineering, University of Toronto, ON, Canada. Previously, he worked as an Associate System Engineer and Data Management Support Specialist with IBM India Private Limited from 2015 to 2018. In 2020, he was a Research Volunteer with the KITE Research Institute, Toronto Rehabilitation Institute, University Health Network, Canada. His research interests focus on unsupervised deep learning, computer vision, multimodal machine learning, and the application of artificial intelligence for detecting behaviours of risk in people with dementia.

He is a recipient of several prestigious awards and grants, including the EPIC-AT Fellowship, the AGE-WELL-UofT FASE Graduate Student Award, and the Wildcat Graduate Scholarship. He serves as an Organizing Committee Member for the Workshop on AI for Aging, Rehabilitation, and Intelligent Assisted Living (at ECML and IJCAI) and as a Program Committee Member for the IJCAI-ECAI Special Track on AI and Health. He is also an active peer reviewer for prominent scientific journals, including IEEE Transactions on Image Processing and IEEE Sensors.
\end{IEEEbiography}

\begin{IEEEbiography}[{\includegraphics[width=1.1in,height=1.25in,clip]{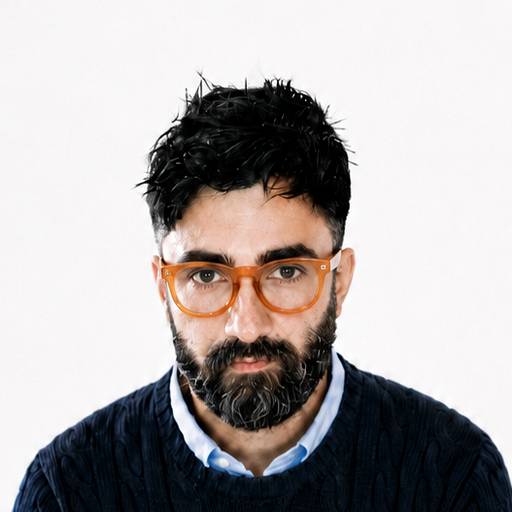}}]{Ali Abedi} is a Machine Learning Specialist at the Peter Munk Cardiac Centre, University Health Network, Canada. He holds a PhD in Electrical Engineering and Computer Science and completed a postdoctoral fellowship at the University of Toronto. His research focuses on artificial intelligence for healthcare, including agentic AI, multimodal and longitudinal clinical data analysis, large language models, cardiovascular imaging, remote patient monitoring, and clinical decision support. He has contributed to more than 40 peer-reviewed publications and has supervised and mentored numerous research trainees.
\end{IEEEbiography}

\begin{IEEEbiography}[{\includegraphics[width=1.1in,height=1.25in,clip]{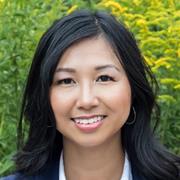}}]{Charlene H. Chu} received the BScN degree (Hons.) and the Ph.D. degree from the University of Toronto, Toronto, ON, Canada, in 2006 and 2016, respectively. She completed a postdoctoral fellowship in occupational science and therapy at the University of Toronto and the KITE Research Institute - Toronto Rehabilitation Institute, University Health Network (UHN), Toronto.

She is currently an Associate Professor with the Lawrence Bloomberg Faculty of Nursing, University of Toronto, where she is cross-appointed to the Institute of Life Course and Aging as well as the Rehabilitation Sciences Institute. She is also an Affiliate Scientist with KITE Toronto Rehab at UHN. Her primary research interests include the design, implementation, and evaluation of technology-enabled interventions to support the health, function, and quality of life of older adults across post-acute care, long-term care, community, and rehabilitation settings.
\end{IEEEbiography}

\begin{IEEEbiography}[{\includegraphics[width=1.1in,height=1.25in,clip]{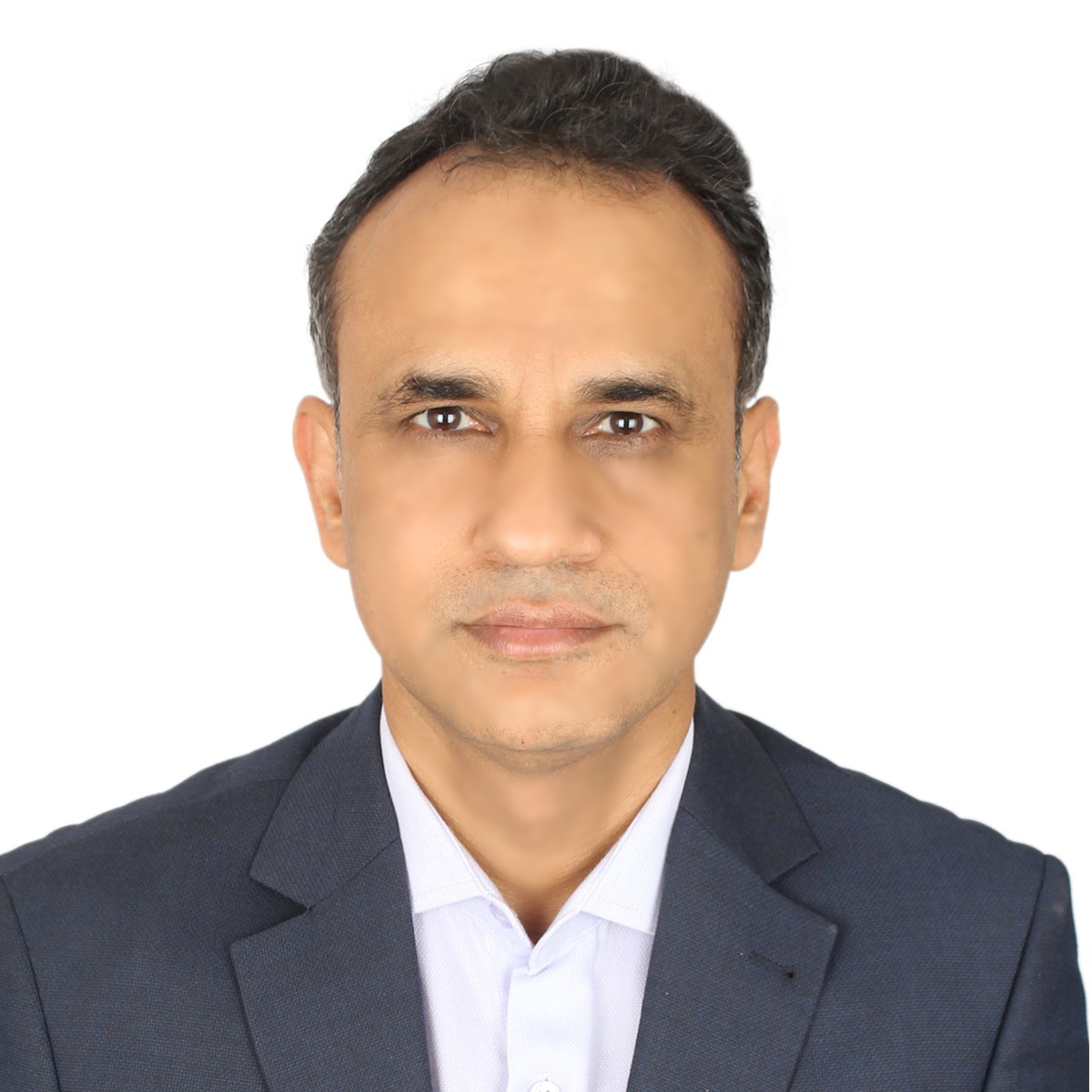}}]{Amir Ahmad} received his Ph.D. degree in computer science from The University of Manchester, Manchester, U.K. He is currently a Professor with the College of Information Technology, United Arab Emirates University, Al Ain, United Arab Emirates. His research interests include machine learning, artificial intelligence, and metamaterials.
\end{IEEEbiography}

\begin{IEEEbiography}[{\includegraphics[width=1.1in,height=1.25in,clip]{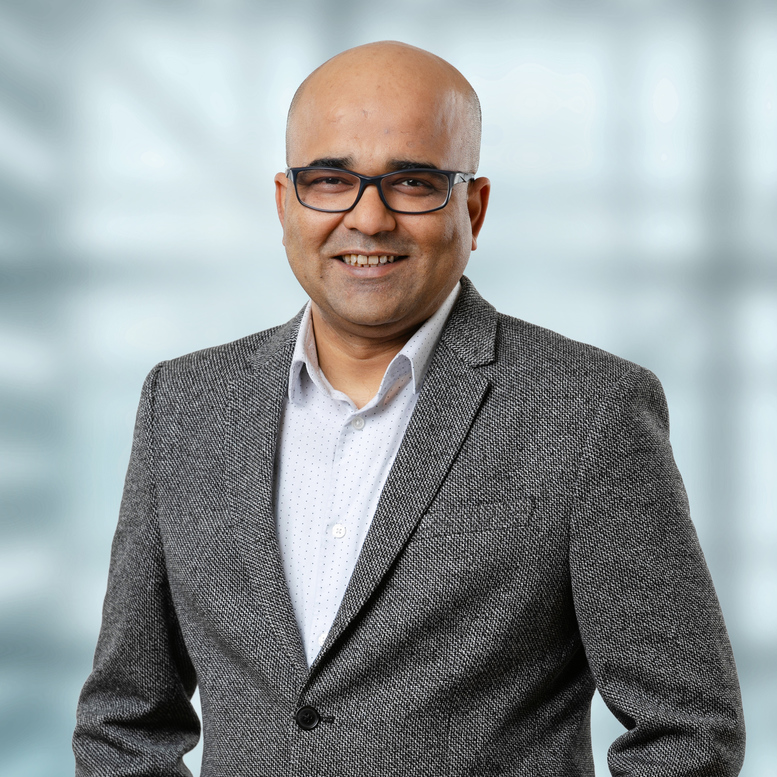}}]{Shehroz S. Khan} is an Assistant Professor at the American University of the Middle East, Kuwait. Previously, he worked as a Scientist at KITE Research Institute, University Health Network, Canada and as an Assistant Professor at the Institute of Biomedical Engineering, University of Toronto, Canada. He holds a PhD Degree from the University of Waterloo, Canada in Computer Science with a specialization in Artificial Intelligence. Dr. Khan’s main research focus is the development of machine learning and deep learning algorithms within the realms of Aging, Rehabilitation and Intelligent Assisted Living. 

As a Principal Investigator (PI) and Co-PI, his research program has been funded through several Canadian, US, other international, and industrial grants. He has published 135 academic papers in top international journals and conferences that have garnered more than 7100 citations (on Google Scholar). He is the founder and organizer of the peer-reviewed International Workshop on Artificial Intelligence for Aging and Rehabilitation that has been held nine times in conjunction with top AI conferences since 2017.
\end{IEEEbiography}

\EOD

\end{document}